\pdfoutput=1
\pdfoutput=1
\pdfoutput=1
\pdfoutput=1
\pdfoutput=1
\documentclass[final, times, numbers, sort&compress]{elsarticle}
\usepackage{lineno,hyperref}
\modulolinenumbers[5]
\journal{Journal of \LaTeX\ Templates}

\usepackage{amsthm}
\usepackage{graphicx}
\usepackage{amsmath}
\usepackage{mathtools}
\usepackage{enumitem}

\usepackage{algpseudocode}
\usepackage{booktabs}
\usepackage[table]{xcolor}
\usepackage{pdflscape}
\usepackage{longtable}
\usepackage{multirow}
\usepackage{geometry}
\usepackage{threeparttablex}
\usepackage{graphicx}
\usepackage[compatibility=false]{caption}
\usepackage[labelformat=simple]{subcaption}
\usepackage[numbers]{natbib}
\usepackage{url}
\usepackage{rotating}

\usepackage{hyperref}
\usepackage{lipsum}
\usepackage{import}
\usepackage[acronym]{glossaries}
\makeglossaries

\usepackage{graphicx}
\usepackage{amsmath}
\usepackage{mathtools}
\usepackage[ruled,linesnumbered,resetcount]{algorithm2e}
\usepackage{algpseudocode}

\usepackage{pdfpages}
\usepackage{import}
\usepackage{multicol}
\usepackage{lineno}
\usepackage{tabularx}

\newcommand{\scalefigure}{0.20}

\newtheorem{definition}{Definition}

\newacronym{ga}{GA}{Genetic Algorithm}
\newacronym{ea}{EA}{Evolutionary Algorithm}
\newacronym{mfo}{MFO}{Multifactorial Optimization}
\newacronym{mfea}{MFEA}{Multifactorial Evolutionary Algorithm}
\newacronym{clustp}{CluSTP}{Clustered Shortest-Path Tree Problem}
\newacronym{sto}{STO}{single-tasking of Genetic Algorithm}
\newacronym{clumrct}{CluMRCT}{Minimum Routing Cost Clustered Tree Problem}
\newacronym{clutsp}{CluTSP}{Clustered Traveling Salesman Problem}
\newacronym{ftsp}{FTSP}{Family Traveling Salesman Problem}
\newacronym{gtsp}{GTSP}{Generalized Traveling Salesman Problem}

\newacronym{tsp}{TSP}{Traveling Salesman Problem}
\newacronym{steinertp}{STP}{Steiner Tree Problem}
\newacronym{clusteinertp}{CluSteinerTP}{Clustered Steiner Tree Problem}
\newacronym{uss}{USS}{Unified Search Space}
\newacronym{de}{DE}{Differential Evolution Algorithm}
\newacronym{pso}{PSO}{Particles Warm Optimization Algorithm}
\newacronym{rmp}{rmp}{Probability of Random Mating}

\newacronym{cesa}{CESA}{}
\newacronym{iim}{IIM}{New Individual Initialization Method}
\newacronym{ncx}{NCX}{New Crossover Operator}
\newacronym{nmo}{NMO}{New mutation Operator}

\newacronym{rpd}{RPD}{Percentage Differences}
\newacronym{nea}{NEA}{New Evolutionary Algorithm}

\begin{document}
\begin{frontmatter}
	
	\title{New Approach for Solving The Clustered Shortest-Path Tree Problem Based on Reducing The Search Space of Evolutionary Algorithm}
	
	\author[httb]{Huynh Thi Thanh Binh}
	
	\author[pdt]{Pham Dinh Thanh\corref{cor1}\href{https://orcid.org/0000-0002-2550-9546 }{\includegraphics[scale=0.5]{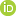}}}
	\ead{thanhpd05@gmail.com}
	
	\author[httb]{Ta Bao Thang}
	
	\cortext[cor1]{Corresponding author.}
	
	\address[httb]{School of Information and Communication Technology, Hanoi University of Science and Technology, Vietnam}
	\address[pdt]{Faculty of Mathematics - Physics - Informatics, Taybac University, Vietnam}
	
	\begin{abstract}
		Along with the development of manufacture and services, the problem of distribution network optimization has been growing in importance, thus receiving much attention from the research community. One of the most recently introduced network optimization problems is the \gls{clustp}. Since the problem is NP-Hard, recent approaches often prefer to use approximation algorithms to solve it, several of which used \glspl{ea} and have been proven to be effective. However, most of the prior studies directly applied \glspl{ea} to the whole \gls{clustp} problem, which leads to a great amount of resource consumption, especially when the problem size is large. To overcome these limitations, this paper suggests a method for reducing the search space of the \glspl{ea} applied to \gls{clustp} by decomposing the original problem into two sub-problems, the solution to only one of which is found by an \gls{ea} and that of the other is found by another method. The goal of the first sub-problem is to determine a spanning tree which connects among the clusters, while the goal of the second sub-problem is to determine the best spanning tree for each cluster. In addition, this paper proposes a new \gls{ea}, which can be applied to solve the first sub-problem and suggests using the Dijkstra's algorithm to solve the second sub-problem. The proposed approach is comprehensively experimented and compared with existing methods. Experimental results prove that our method is more efficient and more importantly, it can obtain results which are close to the optimal results.
	\end{abstract}
	
	
\end{frontmatter}

\glsresetall


\section{Introduction}
\label{Sec_Introduction}
Clustered problems has a wide variety of real-life applications such as optimization of irrigation systems in agriculture, especially irrigation networks in the desert, which has been an urgent problem of mankind. Nowadays, with the development of manufacture and services, economy, optimizations of distribution networks have become increasingly important problems. Therefore, clustered tree problems receive a lot of interest from the research. One of most well-known clustered tree problems is \gls{clusteinertp}, a variant of the Steiner tree problem. \gls{clusteinertp} is applied to applications in various areas~\cite{wu_clustered_2015,dror_generalized_2000,wu2014steiner} such designing transportation or computer networks and design inter-cluster topologies, etc. Recently, the \gls{clustp} has been introduced as theoretical problem in many network optimization problems. As \gls{clustp} is NP-Hard, approximation approaches are often used to solve large instances of \gls{clustp}.

\glspl{ea} are a family of global optimization algorithms that have been proven to be effective in solving many types of problems, both theoretical and practical ones. The mechanism of \glspl{ea} is based on natural selection and Darwinian theory “Survival of the fittest”, which suggest that the quality of individuals in a population should improve after each generation. Recently, the \gls{mfea} has emerged as one of the most effective \glspl{ea} that can be applied to different types of problems. The main concept of the \gls{mfea} is the combination of rules of evolution and cultural transmissions. As a result, the \gls{mfea} has some distinguished features in comparison with classical \glspl{ea}, for example, the \gls{mfea} can solve multiple tasks simultaneously and exploit potential traits of implicit genetic transfer in a multitasking environment.

Accordingly, some research dealing with the \gls{clustp} proposed several evolutionary operators and mechanisms for using the \gls{mfea} to solve the \gls{clustp}. However, because such research often looks for solutions in original search space of the \gls{clustp}, much computational resource and time is consumed. To overcome these drawbacks, this paper proposes a new approach which remodels the \gls{clustp} to a new problem of two parts, one of which is solved by an exact algorithm and the other is solved by an approximation problem, namely a genetic algorithm. Thereby, the part that the genetic algorithm has to deal with has decreased in dimensionality in comparison with the original problem, thus reducing the consumed resource and time. Also, the quality of the ultimate solutions hopefully improves, as a result of the part constructed by the exact algorithm.

The rest of this paper is organized as follows. Section \ref{Notation_and_definitions} presents the notations and definitions used for formulating problem. Section \ref{Sec_Related_Works} introduced related works. The proposed \gls{ea} for the \gls{clustp} is elaborated in section \ref{Sec_Proposed_Algorithm}. Section \ref{Sec_Computational_results} explains the setup of our experiments and analyses the computed results. The paper concludes in section \ref{Sec_Conclusion} with discussions on the future extension of this research.

\section{Notation and definitions}
\label{Notation_and_definitions}
In this paper, a graph is a simple, connected and undirected graph. For a graph $G = (V, E, w)$, $V$ and $E$ are the vertex and the edge sets, respectively, and $w$ is the nonnegative edge length function. An edge between vertices $u$ and $v$ is denoted by $(u, v)$, and its weight is denoted by $w(u, v)$.

For a graph $G$, $V(G)$ and $E(G)$ denote the vertex and the edge sets, respectively. For a vertex subset $U$, the sub-graph of $G$ induced by $U$ is denoted by $G[U]$. For a vertex set $V$, a collection $R = \{R_i | 1 \leq i \leq k\}$ of subsets of $V$ is a partition of $V$ if the subsets are mutually disjoint and their union is exactly $V$. A path of $G$ is simple if no vertex appears more than once on the path. In this paper we consider only simple paths.

For a sub-graph $H$ of $G = (V, E, w)$ and $u, v \in V$, let $d_H(u, v)$ denote the shortest path length between $u$ and $v$ on $H$. Let $d_H(v,U) = \sum_{u \in U} d_H(v, u)$ for a vertex $v$ and a vertex subset $U$. For vertex subsets $U_1$ and $U_2$, let $d_H(U_1,U_2) = \sum_{u \in U_1} d_H(u,U_2)$.

An $st$-path is a path with endpoints $s$ and $t$. Let $P = (v_0, v_1, \ldots , v_p)$ be a $v_0 v_p$ path passing through $v_1, v_2, \ldots , v_{p-1}$ in this order. For $0 \leq i \leq j \leq p$, the sub-path between $v_i$ and $v_j$ of $P$ is denoted by $P[v_i, v_j]$. An edge $(x, y)$ is also thought of as a path. 

\begin{definition}
	For a tree $T$ spanning $S$, i.e., $S \subseteq V(T)$, the local tree of $S$ on $T$ is the sub-tree of $T$ induced by $T[S]$.
\end{definition}

The local tree of $R_i$ in a clustered spanning tree $T$ will be denoted by $L_i(T)$.

\begin{definition}
	Let $R = \{R_i | 1 \leq i \leq k\}$ be a partition of $V$. A spanning tree $T$ is a clustered spanning tree for $R$ if the local trees of all $R_i \in R $ are mutually disjoint, i.e., there exists a cut set $C \subseteq E(T)$ with $|C| = k-1$ such that each component of $T-C$ is a spanning tree $T_i$ for $R_i$ for all $1 \leq i \leq k$. The edges in the cut set $C$ are called inter-cluster edges.
\end{definition}

We call a local tree terminal local tree if it is connected to only one inter-cluster edge. 

The port of a terminal local tree is the vertex adjacent to a vertex not in the cluster.

The \gls{clustp} can be stated as follows:

Given a weighted undirected graph $G = (V, E, w)$ where the vertex set $V$ is partitioned into $k$ clusters ${V_1, V_2, . . ., V_k}$ and the edge set $E$ has a weight function $w: E \to R^+$. One of the vertices $s$ of $G$ is chosen to be the source vertex.


The \gls{clustp} looks for a spanning tree $T$ of $G$ such that:
\begin{enumerate}
	\item For each cluster $V_i (i = 1,\ldots, k)$, a sub-graph including all vertices in $V_i$ is a connected graph.
	\item Minimize $C_T = \displaystyle \sum_{v \in  V} d_{T}(s,v) $.
\end{enumerate}


\section{Related works}
\label{Sec_Related_Works}
Many real life network systems such as irrigation network in agriculture, goods and services distribution systems, cable TV systems and fiber optic systems, can be associated with clustered problems. For their wide range of practical applications, clustered problems have been researched for a long time.

One of the most notable clustered problems is the \gls{clutsp}~\cite{mestria2018new,mestria_grasp_2013,potvin1996clustered}, an important variant of the well-known \gls{tsp}. \gls{clutsp} is applied to applications in various areas~\cite{laporte2002some} such as vehicle routing, manufacturing~\cite{degraeve_optimal_1999}, computer operations~\cite{chisman_clustered_1975, liu_clustering_1999, weintraub_emergency_1999}, examination timetabling~\cite{balakrishnan_scheduling_1992}, cytological testing~\cite{laporte1998tiling}, and integrated circuit testing~\cite{laporte_applications_2002}, etc. \gls{clutsp} was first introduced by Chisman~\cite{chisman_clustered_1975}, who described the problem and presented an optimization model for a warehousing problem. Potvin,~ J.~-Y.~et~al.~\cite{potvin1996clustered} proposed a \gls{ga} to solve the \gls{clutsp} in 1996. The new \gls{ga} used permutation representation, edge recombination crossover~\cite{back_evolutionary_1996}, 2-opt local search heuristic mutation and individual evaluation by transforming the raw fitness of chromosomes using linear ranking. Ding,~C.~et~al.~\cite{ding2007two} applied a new approach based on a Two-Level Evolutionary Algorithm (TLEA) for solving the \gls{clutsp}. In novel approach, the Shortest Hamiltonian cycles in each cluster was determine in the lower level by applying the evolutionary algorithm to the \gls{tsp} problem while in the higher level, a modified evolutionary algorithm with one crossover and two mutation operators is designed to determine the shortest possible solution based on the Hamiltonian cycles generated in the lower level.

One of the newest study about variant of classical \gls{tsp} was done by Pop,~P.~et~al.~\cite{pop2018two}. The authors proposed new approach to solve the \gls{ftsp} by decomposing the problem into two smaller sub-problems: global subproblem and local subproblem. In the global subproblem, the authors use classical \gls{ga} and diploid \gls{ga} for constructing the Hamiltonian tour (called global Hamiltonian tour) connecting among the families. In the local subproblem, the authors look up the visiting order of the required vertices in the families by transforming the global Hamiltonian tour into a \gls{tsp} tour then apply the Concorde algorithm for determining that visiting order.  Applying similar concept in the paper~~\cite{pop2018two},~Pop,~P.~et~al.~\cite{pop2017hybrid} are addressing the \gls{gtsp} but after finding the Hamiltonian tour in global subproblem, a node in a cluster is determined by solving the shortest path problems. Another notable research in solving the \gls{clutsp} was done by Mestria,~M~et~al.~\cite{mestria_grasp_2013}. the authors proposed six heuristic algorithms based on GRASP heuristic for solving the \gls{clutsp}. In the proposed algorithm, the authors also examined several strategies for combining GRASP with path relinking heuristic. 

\glsreset{clusteinertp}

Nowadays, because of the needs for network optimization, clustered tree problems have received a lot of attention. A new variant of the well-known \gls{steinertp}, the \gls{clusteinertp} has also attracted much attention. In the \gls{clusteinertp}, the vertices are divided into groups or clusters. An \gls{steinertp} is a \gls{clusteinertp} if the local tree of each clusters are mutually disjoint~\cite{wu_clustered_2015}.  Wu,~ B.Y.~et~al.~\cite{wu_clustered_2015} showed that the lower and upper bound of Steiner ratios of \gls{clusteinertp} are 3 and 4 respectively. The authors also proposed (2+ $\rho$)-approximation for \gls{clusteinertp} where $\rho$ was the approximation ratio. Lin,~C.-W.~et~al.~\cite{lin_minimum_2016} studied a new variant of clustered tree problem, \gls{clumrct}. The authors showed that the \gls{clumrct} is NP-Hard if the \gls{clumrct} has at least 2 clusters. A 2-approximation algorithm was proposed to solve the \gls{clumrct}. The new algorithm created a two-level graph based on a R-star spanning tree with two characteristics: An R-star spanning tree with minimum routing cost can be generated in $O(n^2)$ time and there exists an R-star whose routing cost is at most twice as much as the optimal cost. The authors also demonstrated that 2-approximation algorithm can solve the Metric \gls{clumrct} in $O(n^2)$ time.

D’Emidio~et~al.~\cite{demidio_clustered_2016} studied the \gls{clustp}, another version of clustered problems. The \gls{clustp} can be found in many real life network optimization problems such as network design, cable TV system and fiber-optic communication. The authors proved that the \gls{clustp} is NP-Hard and proposed an approximation algorithm (hereinafter AAL) for solving the \gls{clustp}. The main idea of AAL is find the minimum spanning tree of each cluster and create a new graph by considering each cluster as a vertex.

As mention above, the approximation approach is more suitable for solving the NP-Hard problems with large dimensionality. Regarding the approximation approach, the \gls{ea} is one of the most effective algorithms for finding the global solution of some problems in different fields~\cite{agoston_eiben_2003,back_evolutionary_1996, goldberg2006genetic}.

\gls{ea} are adaptive search techniques which simulate an evolutionary process like it is seen in nature based on the ideas of the selection of the fittest, crossing and mutation. Thus, \gls{ea} can be used to get approximate solutions for NP-Hard problems. The high adaptability and the generalizing feature of \gls{ea} help to execute these problems by a simple structure.

Recently two new variant of \gls{ea}, \gls{mfea} are applied to solve the \gls{clustp} and improve the resulting solutions.

In~\cite{ThanhPD_TrungTB}, the authors proposed \gls{mfea} (E-MFEA) with new evolutionary operator for finding the solution of \gls{clustp}. The main ideas of the new evolutionary operators is that first constructing spanning tree for smallest sub-graph then the spanning tree for larger sub-graph are construed based on the spanning tree for smaller sub-graph. In~\cite{ThanhPD_DungDA}, the authors take the advantage of Cayley code to encode the solution of \gls{clustp} and proposed evolutionary operators based on Cayley code~\cite{thompson2007dandelion,perfecto2016dandelion,julstrom2005blob,palmer_representing_1994,paulden_recent_2006}. The evolutionary operators are constructed based on ideas of evolutionary operators for binary and permutation representation.

Although experimental results are shown the effective of these algorithms, these algorithm has some shortcomings such as: evolution operators performing on complete graph; finding the solution on large search space, etc.

Therefore, this paper proposes new approach based on the evolutionary algorithm to solve the \gls{clustp}. The new approach improves both running time and qualities of solution.

\section{Proposed Algorithm}
\label{Sec_Proposed_Algorithm}
In this section, we present the novel algorithms for \gls{clustp} problem. The novel algorithm include: new encoding individual for \gls{clustp}, new method for computing the cost of \gls{clustp} and new evolutionary operators such as new initial population, new crossover operator and new mutation operator.
\subsection{Remodeling of the \gls{clustp}} \label{subsec:Remodeling_of_The_CSTP}
Recently, cluster tree problems including the \gls{clustp} have emerged as one of the most interesting research topics. Those problems have received much interest from the academic community and various strategies have been also proposed to produce better solutions. However, most approaches try to search from all valid solutions to find one that has the lowest possible cost. However, when the problem size is great and the problem itself is NP-Hard, like in the case of the \gls{clustp}, it is hard to consider all feasible solutions in a reasonable amount of time.

To overcome this, we propose a new approach which decomposes the \gls{clustp} into two subproblems. The first subproblem, denoted by H-Problem, finds an edge set which connects among the clusters, while the goal of the second subproblem, denoted by L-Problem, is to find a spanning tree for the sub-graph over each cluster. In this approach, spanning trees of sub-graphs over the clusters is constructed after the spanning tree connecting among clusters is found.

\setlength{\intextsep}{0pt}
\renewcommand{\scalefigure}{0.3}
\begin{figure*}[htbp]
	\centering
	\begin{subfigure}{.32\linewidth}
		\centering
		\includegraphics[scale=\scalefigure]{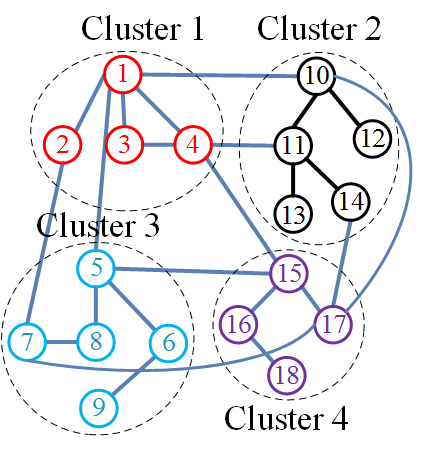}
		\caption{}
		\label{fig:Disadvantage-a}
	\end{subfigure}
	\begin{subfigure}{.32\linewidth}
		\centering
		\includegraphics[scale=\scalefigure]{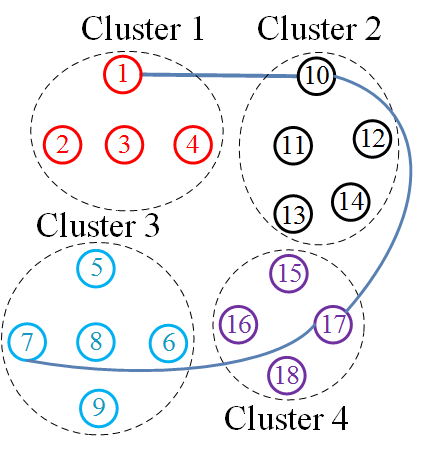}
		\caption{}
		\label{fig:Disadvantage-b}
	\end{subfigure}
	\begin{subfigure}{.32\linewidth}
		\centering
		\includegraphics[scale=\scalefigure]{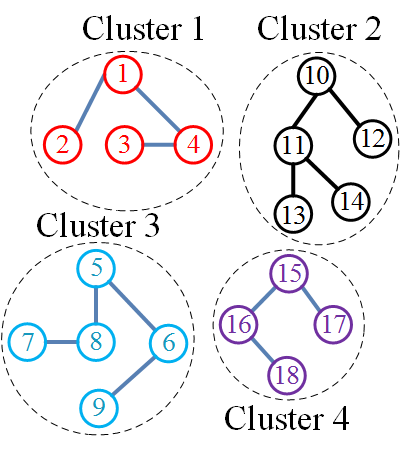}
		\caption{}
		\label{fig:Disadvantage-c}
	\end{subfigure}
	\caption{An example of new approach to solve \gls{clustp}}
	\label{fig:An-example-of-new-approach-to-solve-CSTP}
\end{figure*}
\setlength{\intextsep}{0pt}

Figure~\ref{fig:An-example-of-new-approach-to-solve-CSTP} illustrates an example of the new approach. The input graph is illustrated in Figure~\ref{fig:Disadvantage-a} and Figure~\ref{fig:Disadvantage-b} illustrates a graph which is the solution to the first sub-problem, and Figure~\ref{fig:Disadvantage-c} illustrates spanning trees for sub-graph in clusters.

After the \gls{clustp} is decomposed into two sub-problems, a few existing algorithms such as two-level genetic algorithm \cite{ding2007two,pop2018two,pop2018novel,pop2018twospanningtree} and evolutionary bi-level optimization \cite{deb2014evolutionary,sinha2017evolutionary}  may be applied to solve the subproblems. Although these algorithms use various strategies for improving solution quality as well as reducing resource consumption, in essence, these strategies still deal with all combinations of possible edges connecting all vertices, so the number of cases to consider is still great. In addition, most existing algorithms \cite{ThanhPD_DungDA,ThanhPD_TrungTB} encode solutions to the \gls{clustp} into chromosomes whose numbers of genes are equal to the original dimensionality of input instances, thus also consuming a lot of resource. 

To overcome these drawbacks, we propose a new approach which has two important features:
\begin{itemize}
	\item For each solution to the H-Problem, the novel approach can determine the best corresponding solution to the L-Problem.
	\item To construct a solution to the \gls{clustp}, the novel approach only need to base on a solution of the H-Problem.
\end{itemize}

To achieve the first feature, for each cluster, a solution to the H-Problem stores a vertex that serves as the root of the shortest path tree of the corresponding cluster.  From this vertex, we can use one of the existing exact algorithms such as Dijkstra's algorithm~\cite{shu2012improved,xu2007improved,johnson1973note} for building shortest path tree for the L-Problem. From the above observation, we only need to focus on edges connecting among the vertices of the clusters. Therefore, we only encode the vertices which are nodes of edges connecting among clusters.

For example, with the solution to the H-Problem in Figure~\ref{fig:Disadvantage-b}, our approach stores the edges between vertices 1 and 10; 10 and 17; 17 and 7. Vertices 1, 10, 17 and 7 are used to build the shortest path tree of sub-graphs in the clusters which contain them.

\subsection{New individual representation} \label{subsec:New_individual_representation}
A chromosome is an array of vertex whose i-th element of the array is a vertex which belongs to the i-th cluster. For simplicity, we call the i-th element of chromosome the root of i-th cluster and denote it by $r_i$. The root of the i-th cluster is used for constructing edges which connect the i-th cluster and other clusters.

Note that the root of cluster may be different from the source vertex s.

To construct the edge set of the solution from the set of roots of clusters, we propose a new method \gls{cesa} based on the Dijkstra's Algorithm~\cite{dijkstra1959note} and a feature of graph is created from shortest path tree which is constructed after performing the Dijkstra’s Algorithm. The \gls{cesa} is described in Algorithm~\ref{alg:Construct_Edge_Set_of_Solution}.

\begin{algorithm}[htbp]
	\KwIn{Graph $G=(V,E,C)$ where $C = C_1 \cup C_2 \cup \ldots \cup C_{k}; C_p \cap C_q = \emptyset, \ \forall p \neq q$; Source vertex $s$; An individual  $I=(r_1, r_2, \ldots, r_k)$}
	\KwOut{A Tree $T'=(V', E')$}
	\BlankLine
	\Begin
	{			
		$V' \leftarrow V$\;
		$S \leftarrow \{r_1, r_2, \ldots, r_k\}$\;
		$C_m \leftarrow $ Determine cluster contains $s$\;
		$T \leftarrow $ Determine shortest path tree for $G[S]$ by using Dijstra Algorithm with start vertex $r_m$\;
		
		\ForEach{cluster $C_j$}
		{
			\If{$C_j \neq C_m$}
			{
				$T_j \leftarrow $ Determine shortest path tree for $G[C_j]$ by using Dijstra Algorithm with start vertex $r_j$\;
			}	
			\Else
			{
				$T_m \leftarrow $ Determine shortest path tree for $G[C_m]$ by using Dijstra Algorithm with start vertex $s$\;
			}
		}
		$E' \leftarrow (\cup_{i=1}^{k} E(T_{i})) \cup E(T)$\;
		\Return $T'$
	}
	\caption{Construct Edge Set of Solution}
	\label{alg:Construct_Edge_Set_of_Solution}
\end{algorithm}

Figure~\ref{fig:Representation_individual} illustrates an example of proposed encoding method. Figure~\ref{fig:Representation_a} depicts the input graph G with 6 clusters so number of genes on chromosome is 6. Figure~\ref{fig:Representation_b} illustrates an invalid individual for graph G in which the vertex 3 on clusters 1 is selected randomly as root vertex, vertex 5 on cluster 2 is selected randomly as root vertex, etc. Figure~\ref{fig:Representation_c} presents a \gls{clustp} solution which is constructed from individual in Figure~\ref{fig:Representation_b} by performance the Algorithm~\ref{alg:Construct_Edge_Set_of_Solution}.

\renewcommand{\scalefigure}{0.3}
\begin{figure}[htbp]
	\centering
		\begin{subfigure}[b]{.34\linewidth}
			\centering
			\includegraphics[scale=\scalefigure]{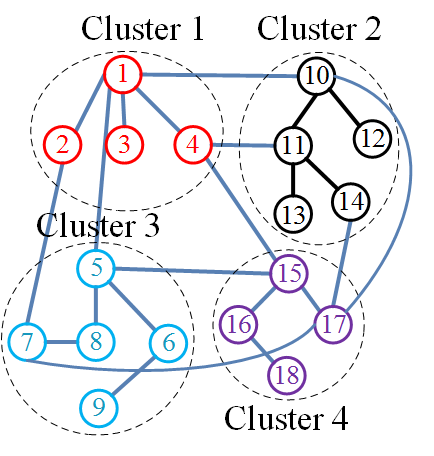}
			\caption{}
			\label{fig:Representation_a}
		\end{subfigure}
		\begin{subfigure}[b]{.28\linewidth}
			\centering
			\includegraphics[scale=\scalefigure]{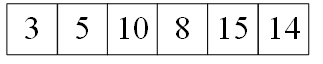}
			\caption{}
			\label{fig:Representation_b}
		\end{subfigure}
		\begin{subfigure}[b]{.34\linewidth}
			\centering
			\includegraphics[scale=\scalefigure]{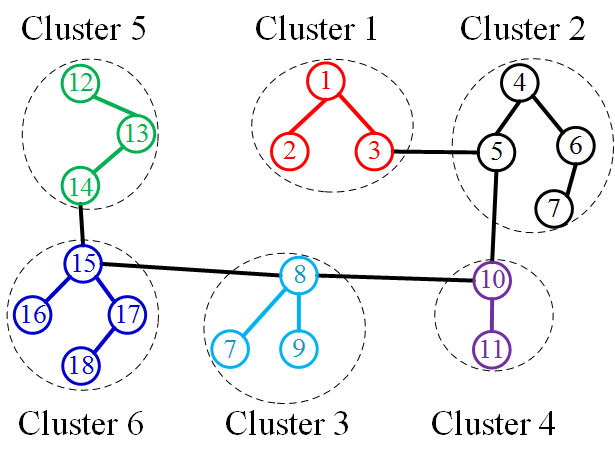}
			\caption{}
			\label{fig:Representation_c}
		\end{subfigure}
	\caption{The representation of individual in unified search space for MFEA with two tasks}
	\label{fig:Representation_individual}
\end{figure}

\subsection{Individual Initialization method} \label{subsec:Individual_Initialization_method}
The \gls{iim} will create randomly an individual\\ Ind=($ind_1, ind_2, \ldots, ind_k$) where $ind_i$ is root of cluster i-th $(i = 1,\ldots, k)$. The detail of \gls{iim} is presented in Algorithm \ref{alg:Create_random_individual}.

\begin{algorithm}[tbp]
	\KwIn{Graph $G=(V,E,C)$ where $C = C_1 \cup C_2 \cup \ldots \cup C_{k}; C_p \cap C_q = \emptyset, \ \forall p \neq q$}
	\KwOut{An individual  $Ind=(ind_1, ind_2, \ldots, ind_k)$}
	\BlankLine
	\Begin
	{			
		$V' \leftarrow \emptyset$\;
		\Repeat{G[V'] is connected graph}
		{
			\For{$i\leftarrow 1$ \KwTo $k$}
			{
				$ind_i \leftarrow$ Select a random vertex in $C_{i}$\;
				$V' \leftarrow  V' \cup \{ind_i\}$\;
			}
		}
		\Return $Ind=(ind_1, ind_2, \ldots, ind_k)$
	}
	\caption{Individual initialization}
	\label{alg:Create_random_individual}
\end{algorithm}

\renewcommand{\scalefigure}{0.36}
\begin{figure}[htbp]
	\centering
	\begin{subfigure}[b]{.32\linewidth}
		\centering
		\includegraphics[scale=\scalefigure]{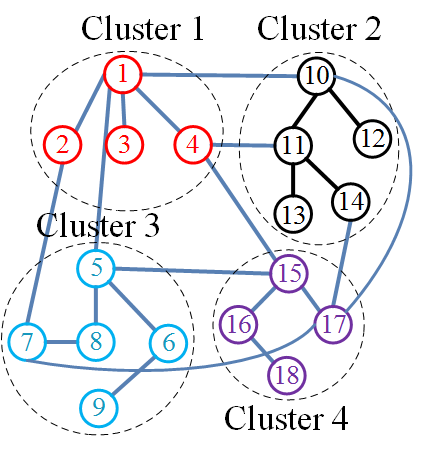}
		\caption{}
		\label{fig:Initial_Method_a}
	\end{subfigure}
	\begin{subfigure}[b]{.32\linewidth}
		\centering
		\includegraphics[scale=\scalefigure]{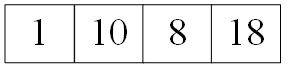}
		\caption{}
		\label{fig:Initial_Method_b}
	\end{subfigure}
	\begin{subfigure}[b]{.32\linewidth}
		\centering
		\includegraphics[scale=\scalefigure]{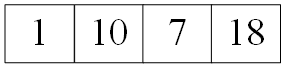}
		\caption{}
		\label{fig:Initial_Method_c}
	\end{subfigure}
	\caption{An example of Individual Initialization method for a graph with 4 clusters}
	\label{fig:Initial_Method}
\end{figure}

Figure~\ref{fig:Initial_Method} illustrates the Individual Initialization method for a graph with 4 clusters. The Figure~\ref{fig:Initial_Method_a} depicts the input graph. The Figure~\ref{fig:Initial_Method_b} depicts an invalid temporary solution because root vertex of cluster 3-th is not connect to other root vertices. This solution will be discard. The Figure~\ref{fig:Initial_Method_c} presents a valid solution.

\subsection{Crossover Operator} \label{subsec:Crossover_Operator}
The \gls{ncx} is proposed based on the two-point crossover operator \cite{back_evolutionary_1996}. However in \gls{ncx}, root of clusters may be not connected so the offspring may be invalid individual. Therefor in \gls{ncx}, if a child is an invalid individual then the child is discarded. The \gls{ncx} is descripted in Algorithm \ref{alg:Proposed_crossover_operator}.

\begin{algorithm}[tbp]
	\KwIn{Graph $G=(V,E,C)$ where $C = C_1 \cup C_2 \cup \ldots \cup C_{k}; C_p \cap C_q = \emptyset, \ \forall p \neq q$;\\
	\qquad \quad \ Two parents: $P_{i}=(p_{i1}, \ldots , p_{ik}), i = 1,2;$}
	\KwOut{Offspring $P^{*}_{i}=(p^{*}_{i1}, \ldots , p^{*}_{ik}), i = 1,2;$}
	\BlankLine
	\Begin
	{	
		$P^{*}_i \leftarrow P_{i}, i = 1,2$\;
		$x_{1,2} \leftarrow$ Select randomly from $\{1,\ldots, k\}$\;
		\lIf{$p_1 > p_2$}
		{
			swap($p_1, p_2$)
		}
		
		\For{$i\leftarrow x_1$ \KwTo $x_2$}
		{
			swap($p^{*}_{1i}, p^{*}_{2i}$)\;
		}
		
		\ForEach{Offspring $P^{*}_i$}
		{		
			\If{$P^{*}_i$ is an invalid individual}
			{
				$P^{*}_i \leftarrow Null $
			}
		}
		
		\Return $P^{*}_{i}$
	}
	\caption{Proposed Crossover Operator}
	\label{alg:Proposed_crossover_operator}
\end{algorithm}

\begin{figure}[htbp]
	\centering
	\begin{subfigure}[b]{.48\linewidth}
		\centering
		\renewcommand{\scalefigure}{0.30}
		\includegraphics[scale=\scalefigure]{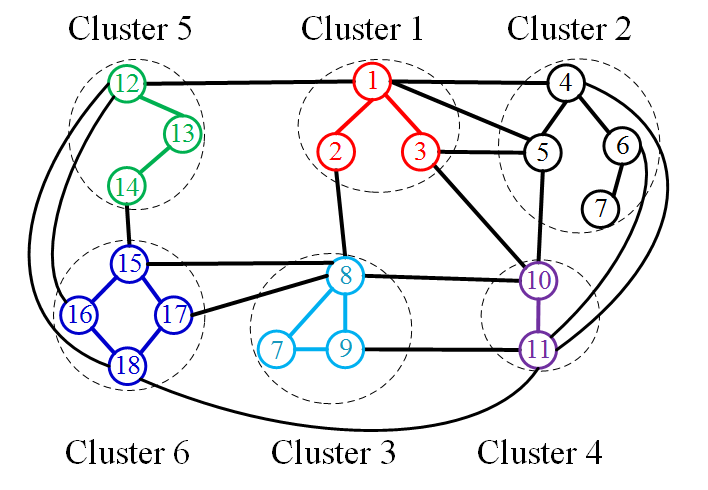}
		\caption{}
		\label{fig:Crossover_Example_a}
	\end{subfigure}
	\begin{subfigure}[b]{.48\linewidth}
		\centering
		\renewcommand{\scalefigure}{0.30}
		\includegraphics[scale=\scalefigure]{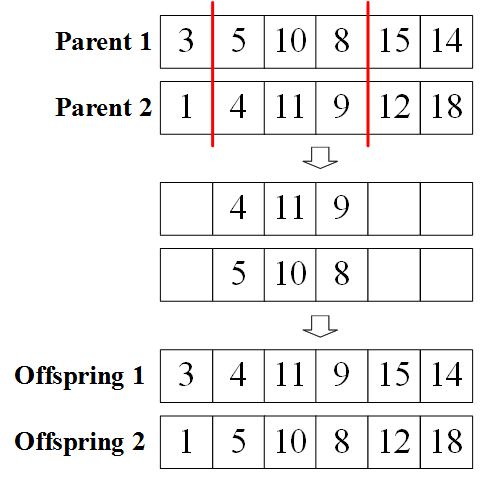}
		\caption{}
		\label{fig:Crossover_Example_b}
	\end{subfigure}
	\caption{An example for the crossover operator}
	\label{fig:Crossover_Example}
\end{figure}

Figure~\ref{fig:Crossover_Example} illustrates the new crossover operator. Figure~\ref{fig:Crossover_Example_a} depicts the input graph G with 6 clusters so the number of genes in chromosome is 6. Figure~\ref{fig:Crossover_Example_b} presents the steps of \gls{ncx} with two vertical red lines illustrate two crossover point. The \gls{ncx} products two offspring. However, the offspring 1 is invalid individual because the sub-graph of G induce by set of vertices on offspring 1 \{3, 4, 11, 9, 15, 14\} is not connected graph so this offspring is discarded.

\subsection{New Mutation Operator} \label{subsec:New_Mutation_Operator}
The \gls{nmo} for the \gls{clustp} will change the root of a cluster. The detail of \gls{nmo} is presented in Algorithm~\ref{alg:Proposed_mutation_operator}.
\\
\begin{algorithm}[tbp]
	\KwIn{Graph $G=(V,E,C)$ where $C = C_1 \cup C_2 \cup \ldots \cup C_{k}; C_p \cap C_q = \emptyset, \ \forall p \neq q$; An individual: $P=(p_{1}, \ldots , p_{k});$}
	\KwOut{An individual $P^{*}=(p^{*}_{1}, \ldots , p^{*}_{k});$}
	\BlankLine
	\Begin
	{   
		$P^{*} \leftarrow P$\;
		Select a random cluster $C_j$\; 
		\Repeat{$P^{*}$ is an invalid individual}
		{			
			\If{All vertices in cluster $C_j$ are considerd}
			{
				\Return $P$;	
			}
			$x \leftarrow$ Select randomly vertex from cluster $C_j$\;
			swap($x, p^{*}_{j}$)\;
		}
		
		\Return $P^{*}$;	
	}
	\caption{Proposed mutation operator}
	\label{alg:Proposed_mutation_operator}
\end{algorithm}

\begin{figure}[htbp]
	\centering
	\begin{subfigure}[b]{.48\linewidth}
		\centering
		\renewcommand{\scalefigure}{0.30}
		\includegraphics[scale=\scalefigure]{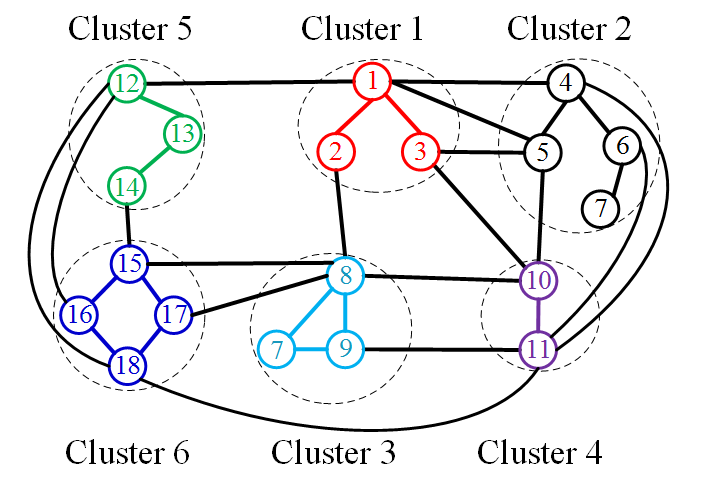}
		\caption{}
		\label{fig:Mutation_Example_a}
	\end{subfigure}
	\begin{subfigure}[b]{.48\linewidth}
		\centering
		\renewcommand{\scalefigure}{0.30}
		\includegraphics[scale=\scalefigure]{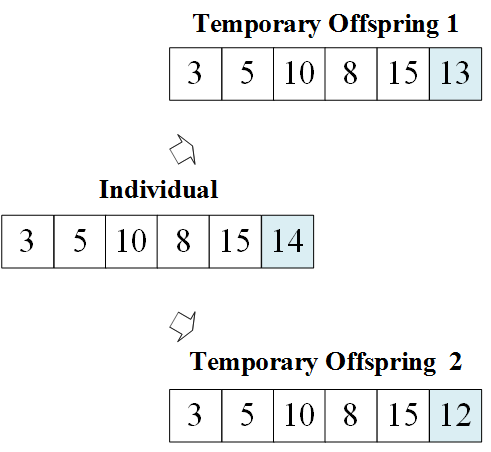}
		\caption{}
		\label{fig:Mutation_Example_b}
	\end{subfigure}
	\caption{An example for the crossover operator}
	\label{fig:Mutation_Example}
\end{figure}
\setlength{\intextsep}{0pt}

Figure~\ref{fig:Mutation_Example} depicts an example of \gls{nmo}. Figure~\ref{fig:Mutation_Example_a} presents the input graph with 6 clusters and 18 vertices. In Figure~\ref{fig:Mutation_Example_a}, the randomly selected cluster is 6-th cluster (cell in blue); temporary offspring 1 obtained by replacing vertex 14 with vertex 13 on genes 6-th on individual but the temporary offspring 1 is an invalid individual so it is discarded; meanwhile, temporary offspring 2 obtained by replacing vertex 14 with vertex 12, this offspring is a valid individual so it is output of \gls{nmo}.

\subsection{New Evaluation Function} \label{subsec:New_Evaluation_function}
In subsection, a new approach for computation the cost of \gls{clustp} individual is proposed. The new approach decreases the consume resource in compared with classical approach.

As mention above, the cost of the solution T is computed as following:

\begin{align}
		f(T) &= \sum_{u \in V} d_T(s,u) \label{eq:New_Evaluation_function_1}\\
			 &= \sum_{i=1}^{k}\sum_{u \in V_i} d_T(s,u)\nonumber \\
			 &= \sum_{i=1}^{k}\sum_{u \in V_i} (d_T(s,r_i) +  d_T(r_i,u))\nonumber \\
			 &= \sum_{i=1}^{k} \left( \left | V_i \right| * (d_T(s,r_i) +  \sum_{u \in V_i} d_T(r_i,u) \right) \label{eq:New_Evaluation_function_2}
\end{align}

The equation~\eqref{eq:New_Evaluation_function_2} point out that the cost of solution T can compute through cost of path from source vertex to root vertices and from root vertices to other vertex in the same cluster with the corresponding root vertex. Because dimensionality of sub-graph is smaller than dimensionality of graph G, resource computation according to equation~\eqref{eq:New_Evaluation_function_2} is reduced in compared with computing according to equation~\eqref{eq:New_Evaluation_function_1}.

\section{Computational results}
\label{Sec_Computational_results}

\subsection{Problem instances}
 Due to the fact that no available instances for the \gls{clustp} were available, we generated a set of test instances based on the MOM \cite{helsgaun_solving_2011} \cite{mestria_grasp_2013} (further on in this paper MOM-lib) of the Clustered Traveling Salesman Problem. The MOM-lib included six distinct types of instances which were created through various algorithms \cite{demidio_clustered_2016} and categorized into two kinds according to dimensionality: small instances, each of which had between 30 and 120 vertices and large instances, each of which had over 262 vertices. The instances were suitable for evaluating cluster problems \cite{mestria_grasp_2013}.
 
 However, in order to test the proposed algorithms' effectiveness in solving the \gls{clustp}, it was necessary to add information about a source vertex to each of the instances. Therefore, we selected a random vertex as the source vertex for each instance.
 
 For evaluation of the proposed algorithms, instances with dimensionality from 30 to 500 were selected.
 
 
 All problem instances are available via \cite{Pham_Dinh_Thanh_2018_Instances} 

 \subsection{Experimental setup}
 To evaluate the performance of new \gls{ea} for the \gls{clustp}, we implemented two sets of experiments.
 \begin{itemize}
 	\item[$\bullet$] On the first set, the approximation algorithm C-MFEA~\cite{ThanhPD_DungDA} and E-MFEA~\cite{ThanhPD_TrungTB} were implemented. Then the results of these algorithms for each instance were compared with those by \gls{nea} in terms of solution quality and run times.
 	\item[$\bullet$] On the second set, since the performance of the proposed~\gls{ea} were contributed by parameter: number of clusters in the \gls{clustp} instance and the average number of vertices in a cluster, the convergence trends of each task in generations, we conducted experiments for evaluating the effect of these parameters. 
 \end{itemize}
 
 Each scenario was simulated for 30 times on the computer (Intel Core i7 - 4790 - 3.60GHz, 16GB RAM), with a population size of 100 individuals evolving through 500 generations which means the total numbers of task evaluations are 50000, the random mating probability is 0.5 and the mutation rate is 0.05. The source codes were installed by Visual C\# language.

 \subsection{Experimental criteria}
 We focuses on the following criteria to assess the quality of the output of the algorithms.
 
 \noindent

\begin{table}[htb] 
	 \centering
	\begin{tabular}{p{4.5cm} p{9.3cm}}
		\hline 
		\multicolumn{2}{c}{\textbf{Criteria}} \\  	
		\hline 
		Average (Avg) & The average function value over all runs \\ 
		\hline 
		Coefficient of variation (CV) & Ratio of standard deviation to the averaged function value \\ 
		\hline 
		Best-found (BF) & Best function value achieved over all runs \\ 
		\hline 
	\end{tabular}
	\label{tab:Critera}
\end{table}

To compare the performances of two algorithms, we compute the Relative Percentage Differences (RPD) between the average results obtained by the algorithms. The RPD is computed by the following formula:
\[
	RPD(A) = \dfrac{C_A - C_{New}}{C_{New}}*100\%
\]
where $C_{New}$ is average cost of the solutions obtained by the proposed algorithm and $C_A$ is average cost of the solutions obtained by one of the compared existing algorithms, which are C-MFEA, E-MFEA and AAL.

We also compute the gap between the costs of the results obtained by algorithm A and B with the following formula:
\[
PI(A,B)=\dfrac{C_B - C_A}{C_B}*100\%
\]
where $C_A$, $C_B$ is the cost of the best solution generated by algorithm A, B respectively.

\subsection{Experimental Results on Euclidean Instances}
\subsubsection{Comparison of the Performances of AAL, C-MFEA, E-MFEA and The Novel Evolutionary Algorithm}
A criterion in comparing the results obtained by the proposed algorithm and two existing algorithms is running time. The execution time of the \gls{nea} is shorter than that of C-MFEA (at least 2 times) and that of E-MFEA (at least 5 times). The reason behind this is that \gls{nea} reduces the dimensionalities of the input problem so the genetic operators only perform in chromosome having dimensionality equal to the number of clusters. Due to the number of clusters is much smaller than number of vertices, the running time of \gls{nea} is greatly decreased.

In terms of cost, the \gls{nea} outperform than two existing algorithms on all instances. The average of PI(\gls{nea}, E-MFEA) and of PI(\gls{nea}, C-MFEA) are 35.1\% and 65.8\% respectively. However, the PI(\gls{nea}, E-MFEA) and PI(\gls{nea}, C-MFEA) on instances in each type are distinct. The PI() on small instances are two-thirds of PI() on large instances, which mean that the \gls{nea} is more effective than two existing algorithms when dimensionality of instances is larger. The reason for these results is that the dimensionality of input problem when using E-MFEA and C-MFEA algorithm (equal to number vertices) increases more than that of input problem using \gls{nea} algorithm (equal to the number of clusters). Therefore, in comparison with the optimal solution, the accuracy of the solutions obtained by C-MFEA and E-MFEA decreases faster than that obtained by \gls{nea}. The details of PI() of each type are presented in the Table \ref{tab:Summary-of-Results-Obtained-By-3-algorithms}.

\begin{table}[htbp]
	\centering
	\caption{Summary of Results Obtained By E-MFEA, C-MFEA and \gls{nea} on Instances}
	\begin{tabular}{|l|c|c|}
		\hline
		& \multicolumn{1}{l|}{\textbf{PI(\gls{nea}, E-MFEA)}} & \multicolumn{1}{l|}{\textbf{PI(\gls{nea}, C-MFEA)}} \\
		\hline
		Type 1 Small & 20.70\% & 50.30\% \\
		\hline
		Type 1 Large & 33.50\% & 75.80\% \\
		\hline
		Type 3 & 37.70\% & 65.50\% \\
		\hline
		Type 4 & 64.80\% & 85.90\% \\
		\hline
		Type 5 Small & 25.80\% & 48.10\% \\
		\hline
		Type 5 Large & 36.70\% & 70.40\% \\
		\hline
		Type 6 Small & 22.70\% & 54.70\% \\
		\hline
		Type 6 Large & 38.90\% & 76.00\% \\
		\hline
	\end{tabular}%
	\label{tab:Summary-of-Results-Obtained-By-3-algorithms}%
\end{table}%

More details of the results obtained by these algorithms are provided in Table~\ref{tab:ResultsType1}, Table~\ref{tab:ResultsType34}, Table~\ref{tab:ResultsType5} and Table~\ref{tab:ResultsType6}. Using the Cayley code~\cite{thompson2007dandelion,perfecto2016dandelion,julstrom2005blob,palmer_representing_1994,paulden_recent_2006}, which can be used in encoding spanning trees of graphs with more than two nodes, the C-MFEA is not be applied to \gls{clustp} problem instances with less than three clusters. Hence, in Table~\ref{tab:ResultsType1}, Table~\ref{tab:ResultsType34}, Table~\ref{tab:ResultsType5} and Table~\ref{tab:ResultsType6}, symbol "-" indicates that the corresponding problem instances could not be solved by the C-MFEA.


\subsubsection{Analysis of influential factors}

It is clear that the cost of the solution to an instance depends on the total cost of the edges. However, as the \gls{clustp} problem is NP-Hard, it is too much work to check all possible combinations of edges in a problem instance. Moreover, problem instances are created by various algorithms, so it takes a lot of time and effort to find the trend of the performance of the new algorithm based on the weight of the edges. Therefore, we want to determine other factors which influence the performance of the new algorithm and deduce the trend of the results obtained by the algorithm.

The chromosome in the new algorithm is influenced by the number of clusters in the \gls{clustp} instance, so the number of clusters affects the efficiency of the evolutionary operators as well as the performance of the novel algorithm. Another possible factor is the average number of vertices in a cluster which does not directly appear in the evolutionary algorithm but affects the solution through the spanning tree obtained by the Dijsktra's algorithm from the local root of each cluster.

Therefore, we will analyze the number of clusters and the average number of vertices in a cluster for finding the tendency of the results obtained by the algorithm.

As the number of instances that can be solved by the C-MFEA in each types is small, we only examine the impact of influential factors on the newly proposed evolutionary algorithm and the E-MFEA.

To see the impact of the number of clusters and the average number of vertices in a cluster on the RPD, first we graph scatter plots of the relationship between number of cluster, average number of vertices in a cluster and the RPD for each type, and then we try to find the correlation coefficient in that relationship.

\setlength{\intextsep}{0pt}
\begin{figure}[htbp]
	\centering
	\begin{subfigure}[b]{.48\linewidth}
		\centering
		\renewcommand{\scalefigure}{0.55}
		\includegraphics[scale=\scalefigure]{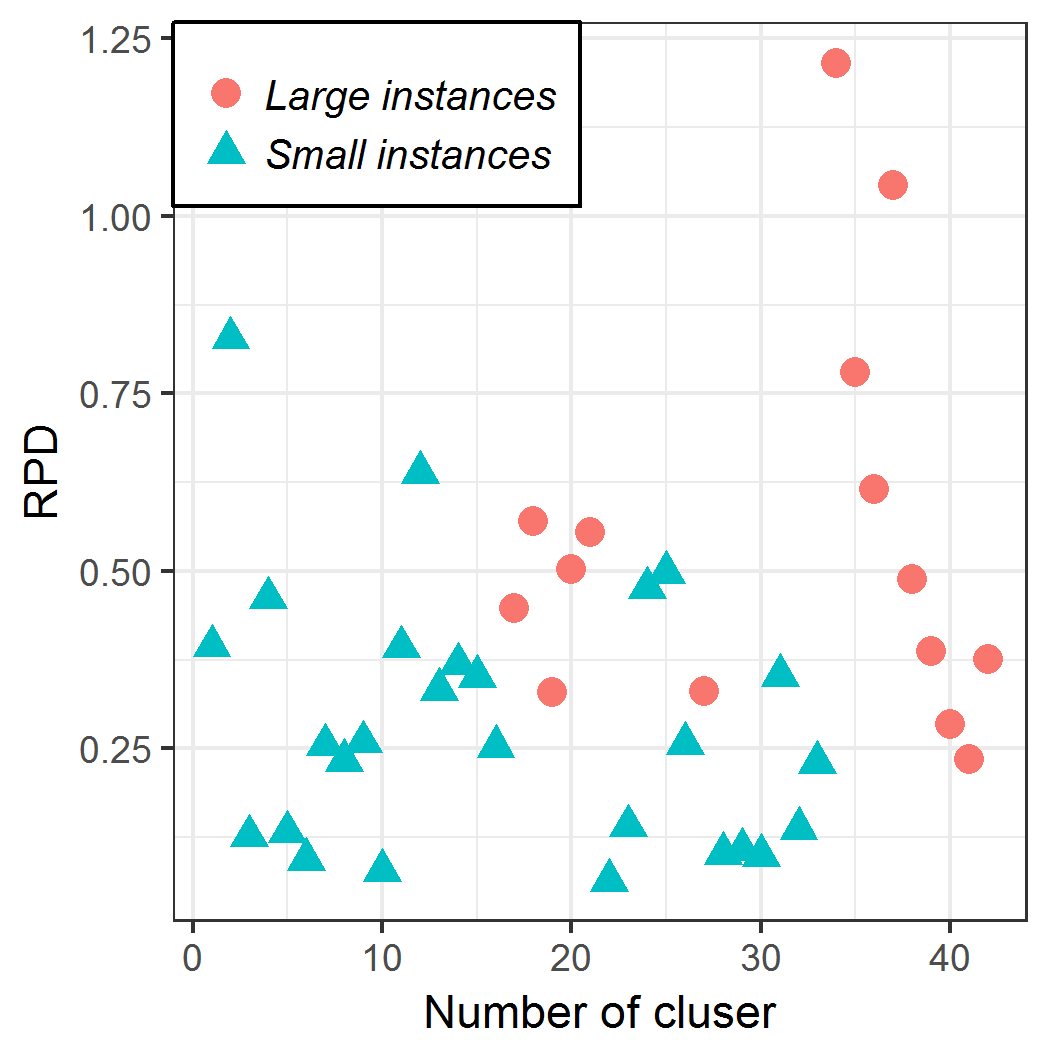}
		\caption{Type 1}
		\label{fig:RPD-NumCluster-Type1}
	\end{subfigure}
	\begin{subfigure}[b]{.48\linewidth}
		\centering
		\renewcommand{\scalefigure}{0.55}
		\includegraphics[scale=\scalefigure]{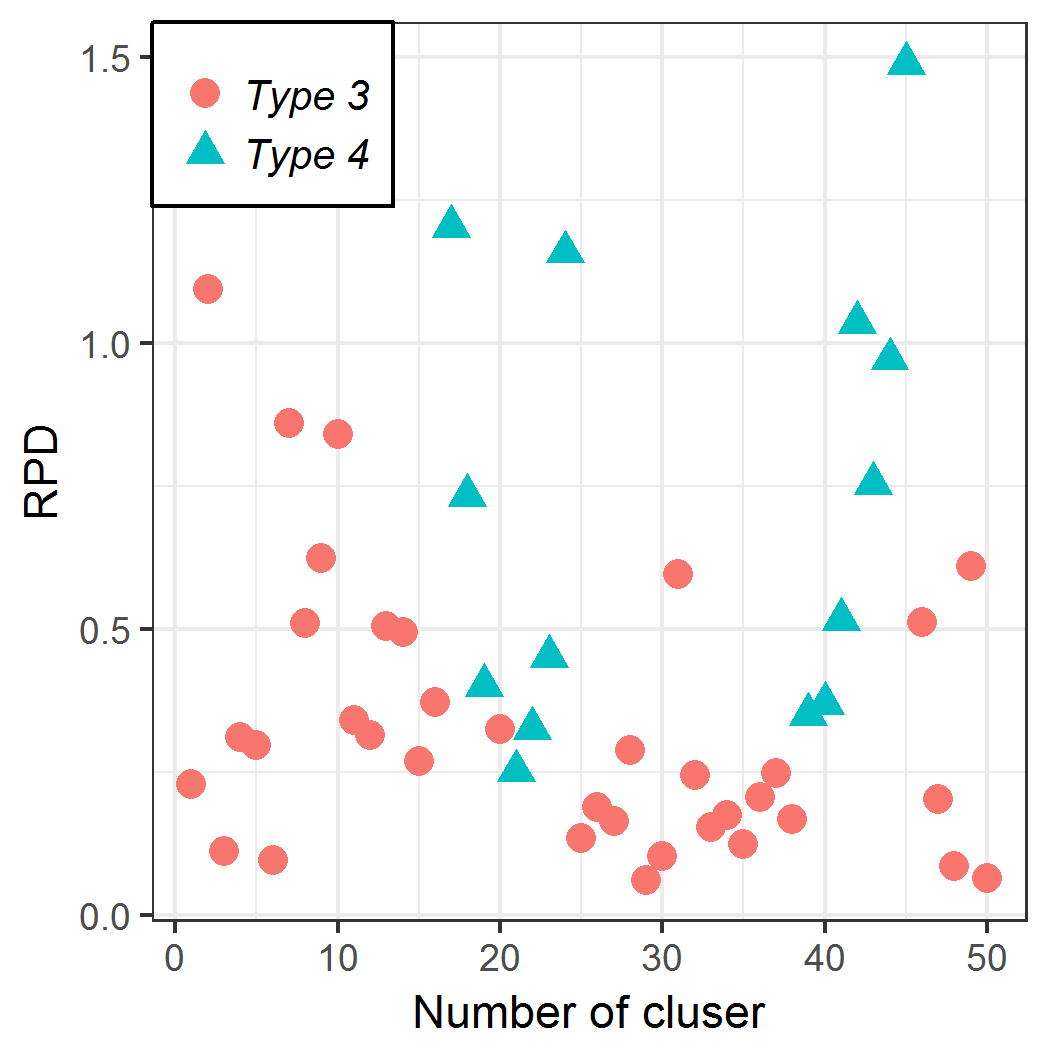}
		\caption{Type 3 and Type 4}
		\label{fig:RPD-NumCluster-Type34}
	\end{subfigure}
		\begin{subfigure}[b]{.48\linewidth}
		\centering
		\renewcommand{\scalefigure}{0.55}
		\includegraphics[scale=\scalefigure]{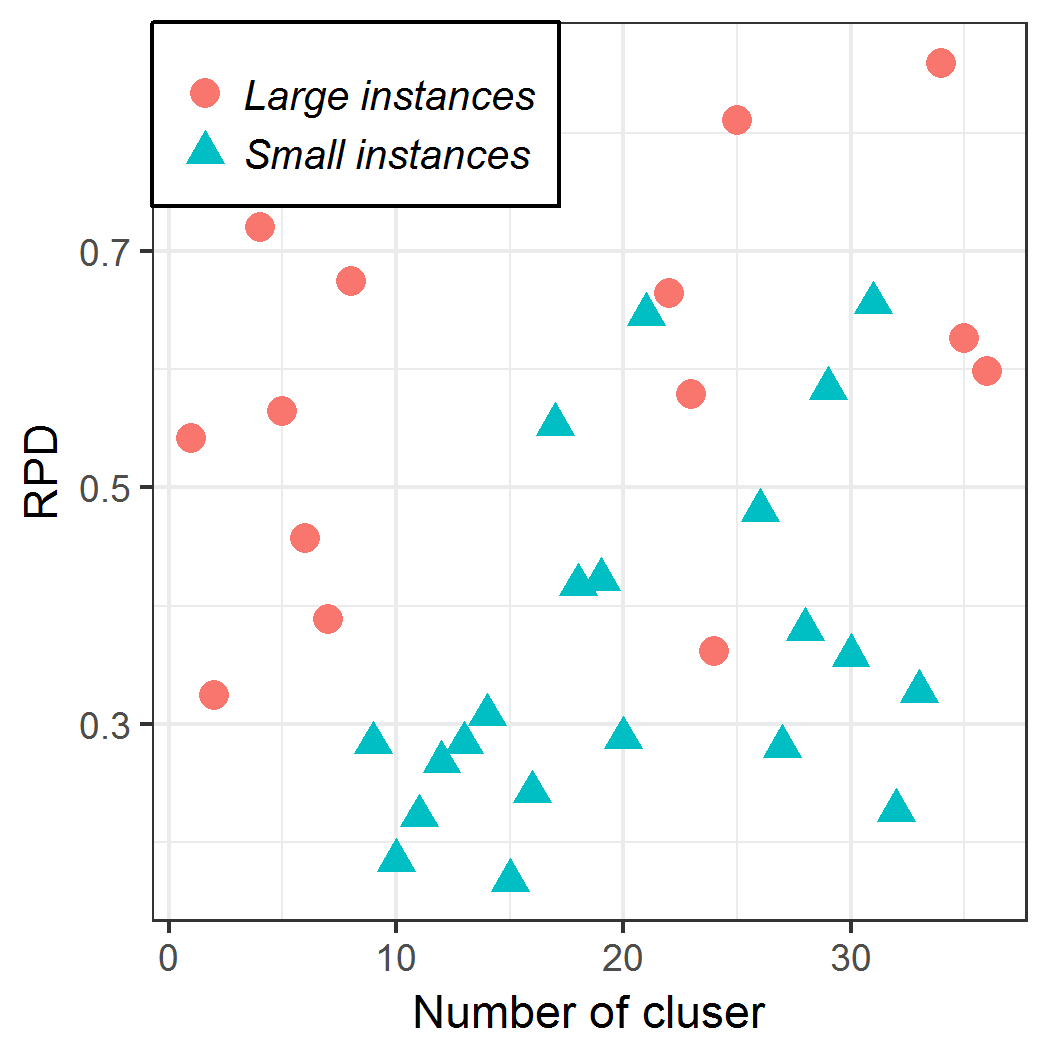}
		\caption{Type 5}
		\label{fig:RPD-NumCluster-Type5}
	\end{subfigure}
	\begin{subfigure}[b]{.48\linewidth}
		\centering
		\renewcommand{\scalefigure}{0.55}
		\includegraphics[scale=\scalefigure]{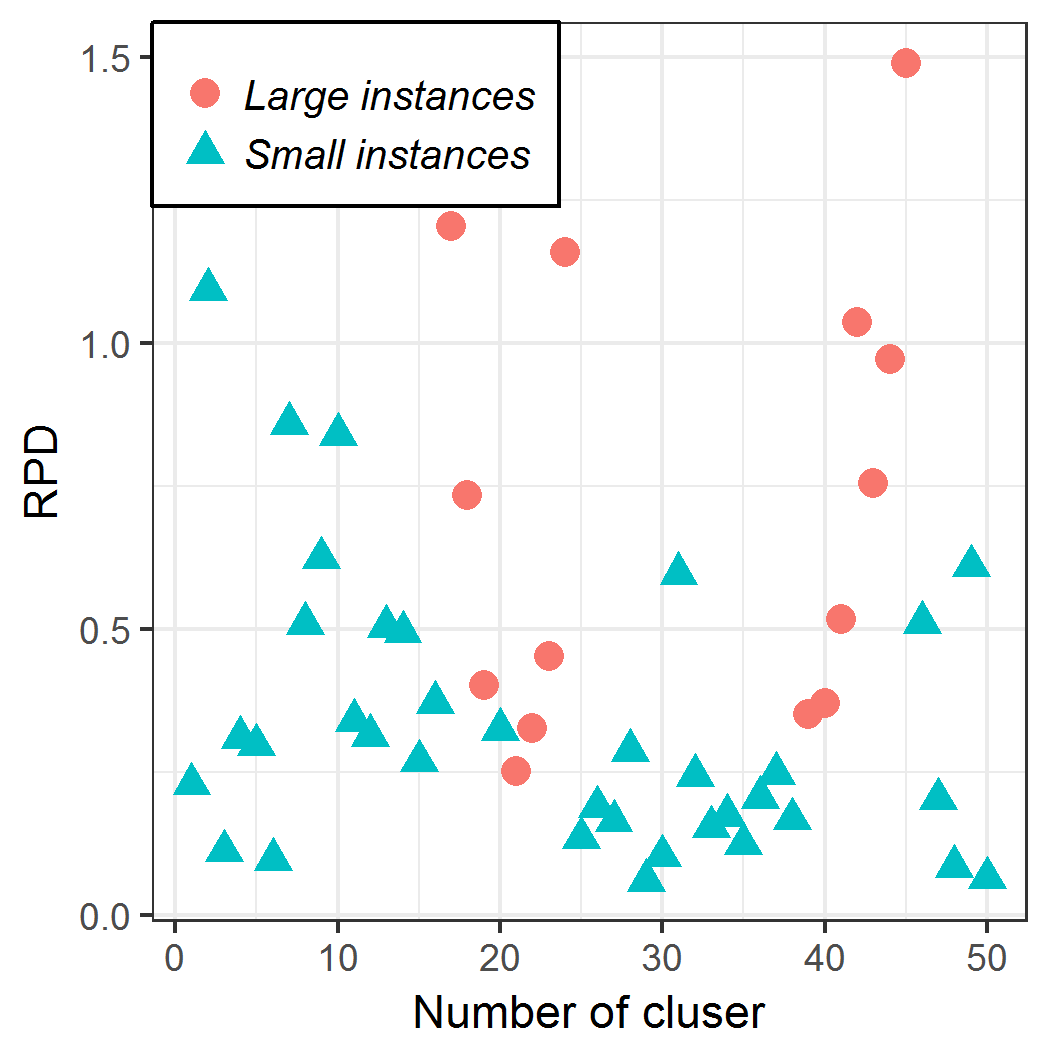}
		\caption{Type 6}
		\label{fig:RPD-NumCluster-Type6}
	\end{subfigure}
	\caption{The scatter of RPD and number cluster in types}
	\label{fig:Scatter-of-RPD-and-number-cluster}
\end{figure}
\setlength{\intextsep}{0pt}

We construct the scatter plots of the number of clusters and the RPD(E-MFEA) as shown in Figure~\ref{fig:Scatter-of-RPD-and-number-cluster}. As can be seen from Figure~\ref{fig:Scatter-of-RPD-and-number-cluster}, it is hard to determine the relationship between the number of clusters and the RPD(E-MFEA) for each type. However, some characteristics of the relationship can still be seen. On instances of Type 3 (Figure~\ref{fig:RPD-NumCluster-Type34}) and small instances of Type 6 (Figure~\ref{fig:RPD-NumCluster-Type6}), the RPD tend to decrease when the number of clusters increases. In other words, the new algorithm is more effective on instances with smaller number of clusters. On the contrary, on small instances in Type 5 (Figure~\ref{fig:RPD-NumCluster-Type5}), the performance of the new algorithm tends to increase when the number of clusters increases.

\setlength{\intextsep}{0pt}
\begin{figure}[htbp]
	\centering
	\begin{subfigure}[b]{.48\linewidth}
		\centering
		\renewcommand{\scalefigure}{0.55}
		\includegraphics[scale=\scalefigure]{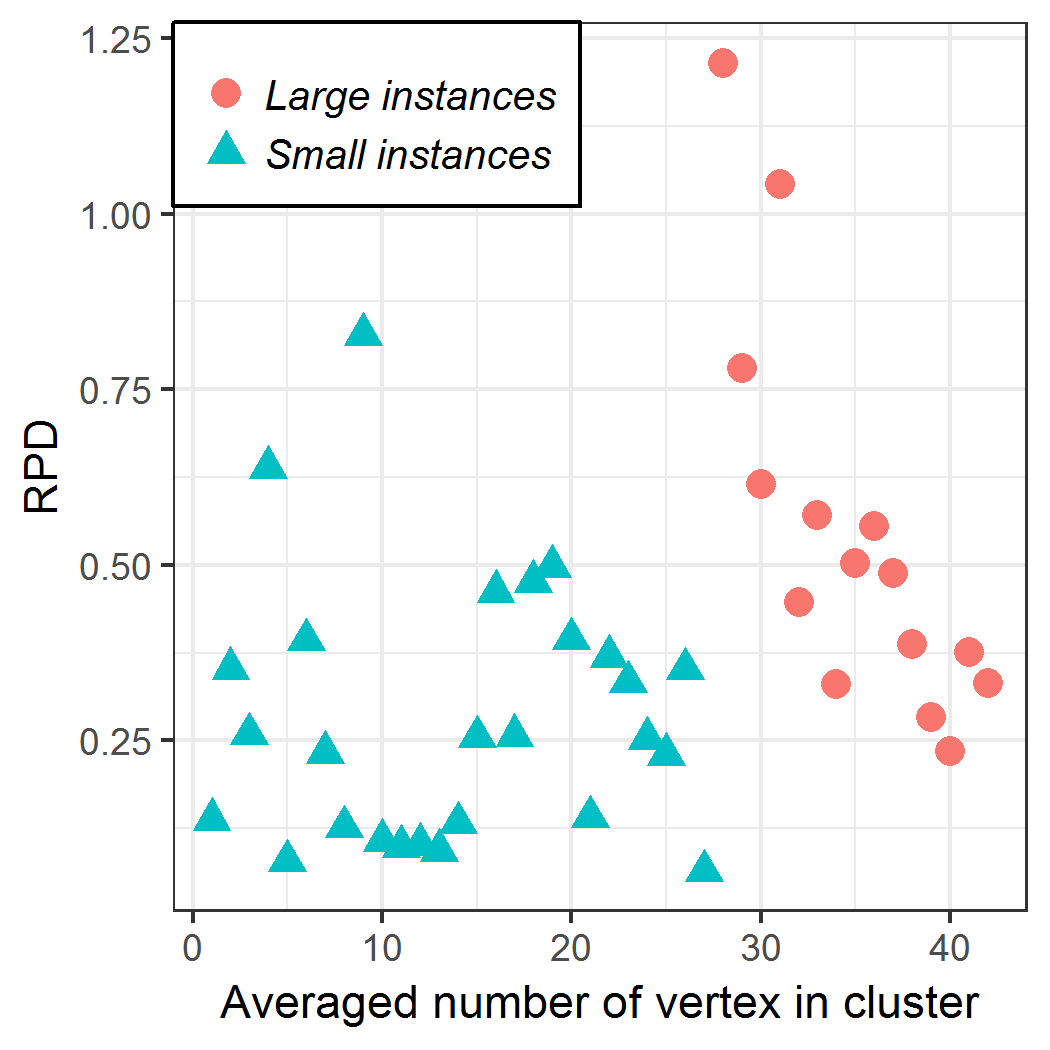}
		\caption{Type 1}
		\label{fig:RPD-AveragedVertex-Type1}
	\end{subfigure}
	\begin{subfigure}[b]{.48\linewidth}
		\centering
		\renewcommand{\scalefigure}{0.55}
		\includegraphics[scale=\scalefigure]{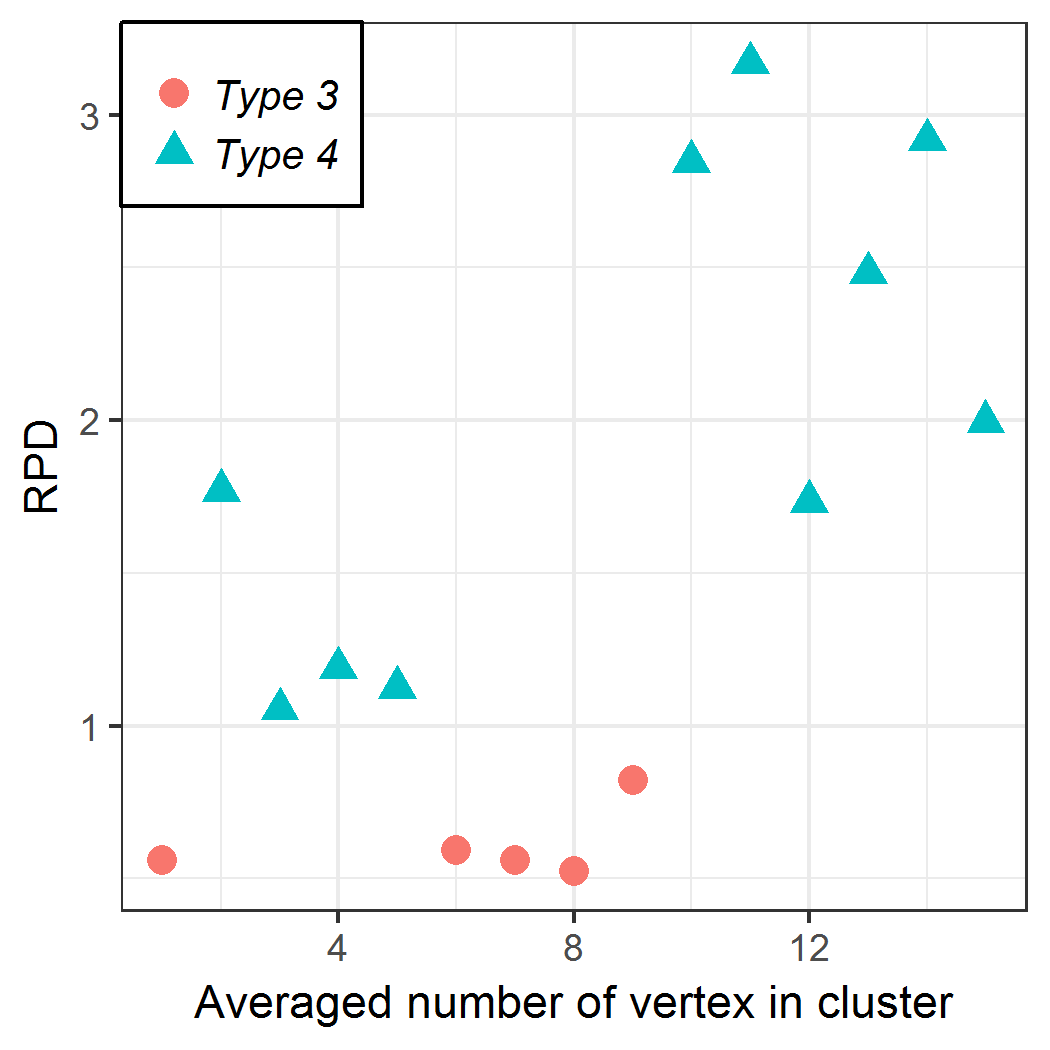}
		\caption{Type 3 and Type 4}
		\label{fig:RPD-AveragedVertex-Type34}
	\end{subfigure}
	\begin{subfigure}[b]{.48\linewidth}
		\centering
		\renewcommand{\scalefigure}{0.55}
		\includegraphics[scale=\scalefigure]{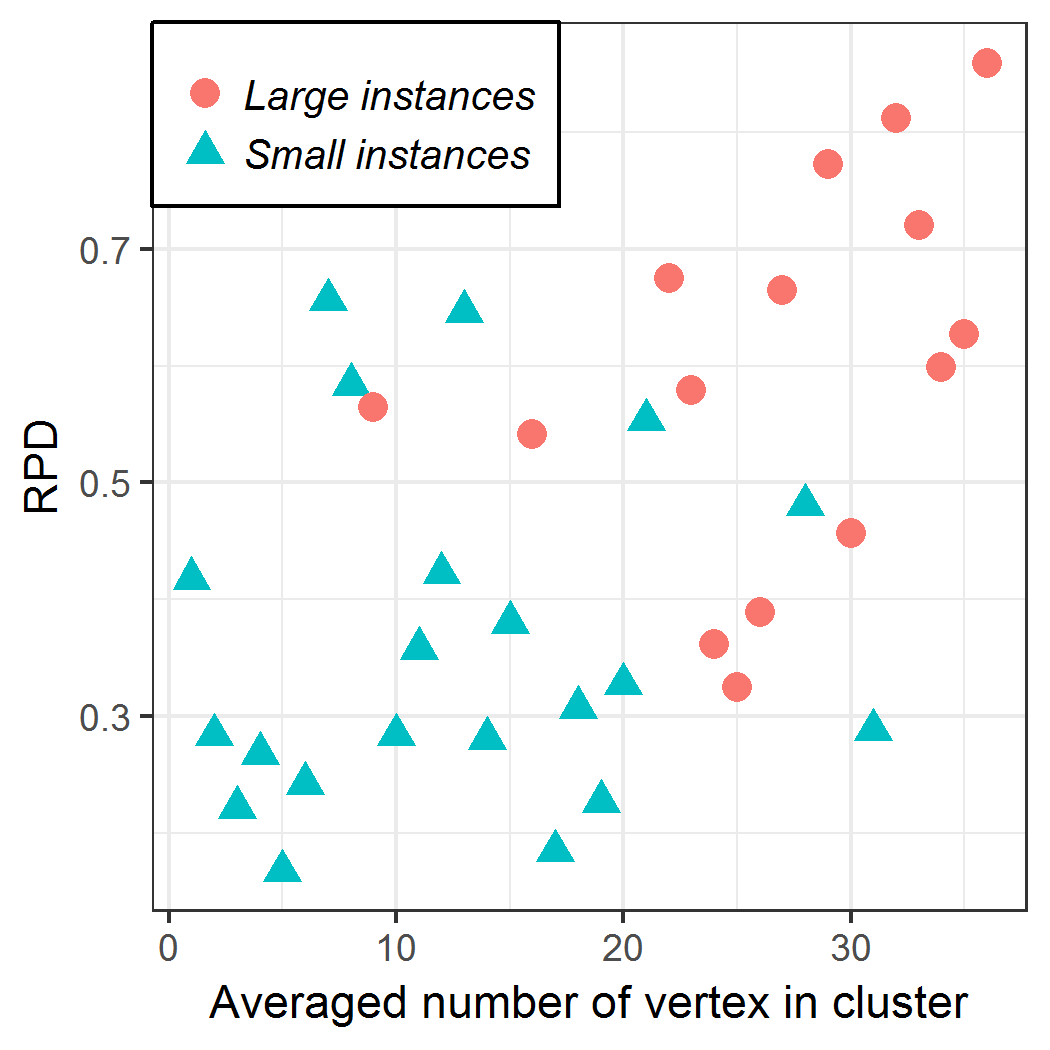}
		\caption{Type 5}
		\label{fig:RRPD-AveragedVertex-Type5}
	\end{subfigure}
	\begin{subfigure}[b]{.48\linewidth}
		\centering
		\renewcommand{\scalefigure}{0.55}
		\includegraphics[scale=\scalefigure]{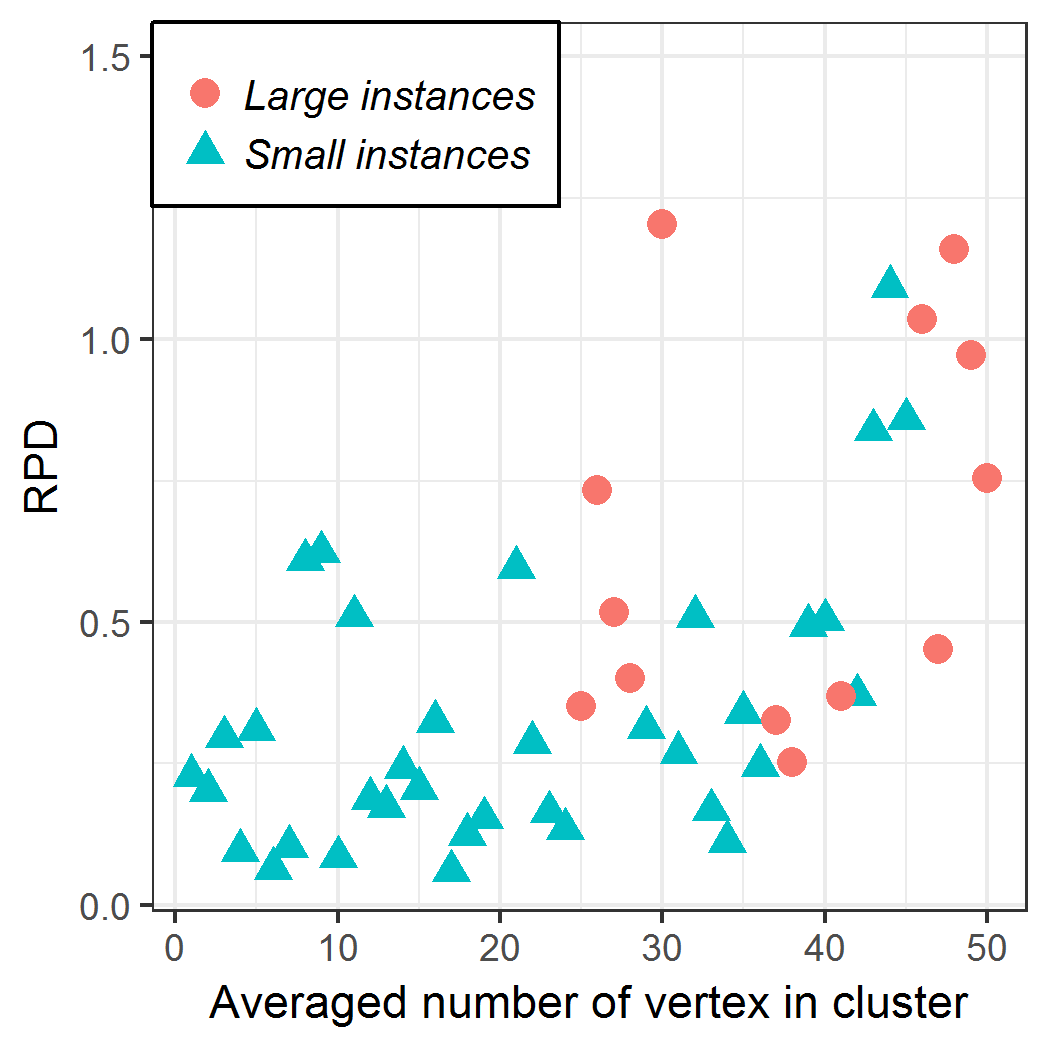}
		\caption{Type 6}
		\label{fig:RPD-AveragedVertex-Type6}
	\end{subfigure}
	\caption{The scatter of RPD and averaged number of vertex in cluster}
	\label{fig:Scatter-of-RPD-and-averaged-number-of-vertex}
\end{figure}
\setlength{\intextsep}{0pt}

The scatter plots of the RPD(E-MFEA) and the average number of vertices in a cluster are presented in Figure~\ref{fig:Scatter-of-RPD-and-averaged-number-of-vertex}. The figures show that in Type 1, Type 4, Type 5 and Type 6, the RPD(E-MFEA) tends to increase when the average number of vertices in a cluster increases. However, the trend of the relationship between RPD(E-MFEA) and the average number of vertices in a cluster varies among types. The tendency of the relationship in Type 1 is not as obvious as those in Type 5 and Type 6. The results in Figure~\ref{fig:RPD-AveragedVertex-Type34} indicate that in type 3, the average number of vertices has hardly any impact on the RPD(E-MFEA).

From the above analysis, it can be concluded that on problem instances in Type 3, the average number of vertex does not have a significant impact on the performance of the new algorithm. In contrast, for the remaining types, the performance of the new algorithm tends to improve as the average number of vertices in a cluster increases.

\begin{table}[htbp]
	\centering
	\caption{Pearson correlation between the number of clusters, the average number of vertices and RPD on the datasets}
	\begin{tabular}{|c|l|r|r|}
		\hline
		\multicolumn{1}{|l|}{Type} & Criteria & \multicolumn{1}{l|}{Pearson pmc} & \multicolumn{1}{l|}{p-values} \\
		\hline
		\multirow{2}[4]{*}{Type 1} & Averaged number of vertex & 0.6819712 & 6.59E-07 \\
		\cline{2-4}          & Number of cluster & -0.3147793 & 0.04232 \\
		\hline
		\multirow{2}[4]{*}{Type 5} & Averaged number of vertex & 0.5610308 & 0.0003716 \\
		\cline{2-4}          & Number of cluster & 0.2280425 & 0.181 \\
		\hline
		\multirow{2}[4]{*}{Type 6} & Averaged number of vertex & 0.8478091 & 8.03E-15 \\
		\cline{2-4}          & Number of cluster & -0.3514797 & 0.01232 \\
		\hline
	\end{tabular}%
	\label{tab:Pearson-Correlation}%
\end{table}%

So far we have examined the relationship among the average number of vertices, the number of clusters and RPD() as well as the impact of the average number of vertices and the number of clusters on the performance of the proposed algorithm. We will take a closer look at the influence of average number of vertex and number of cluster on the performance of the new algorithm.

To analyze more closely the relationship between the number of cluster, average number of vertex in cluster and RPD(E-MFEA), we compute Pearson correlation coefficient (Pearson pmc) of these factors. Because the numbers of instances in Type 3 and Type 4 are very small, we only consider the instances of Type 1, Type 5 and Type 6.

Table~\ref{tab:Pearson-Correlation} presents details of the Pearson pmc of the number of clusters, the average number of vertices in a cluster and the RPD(E-MFEA). In Table~\ref{tab:Pearson-Correlation}, the Pearson pmc values between the average number of vertices and the RPD is larger than the Pearson pmc value between the number of clusters and the RPD. Additionally, in Type 5, correlation coefficient between the number of clusters and the RPD is not statistically significant (p-value = 0.181), which means that the relationship between the average number of vertices and the RPD is stronger than the relationship between the number of clusters and the RPD.

Another noteworthy point in the results in Table~\ref{tab:Pearson-Correlation} is that the values of the Pearson pmc between the number of clusters and the RPD in Type 1 and Type  6 are negative, which means that when the number of clusters increases, the RPD tends to decrease. This statement is reinforced by the results illustrated in Figure~\ref{fig:Scatter-of-RPD-and-number-cluster}.

%

\setlength{\intextsep}{0pt}
\begin{figure}[htbp]
	\centering
	\begin{subfigure}[b]{.48\linewidth}
		\centering
		\renewcommand{\scalefigure}{0.55}
		\includegraphics[scale=\scalefigure]{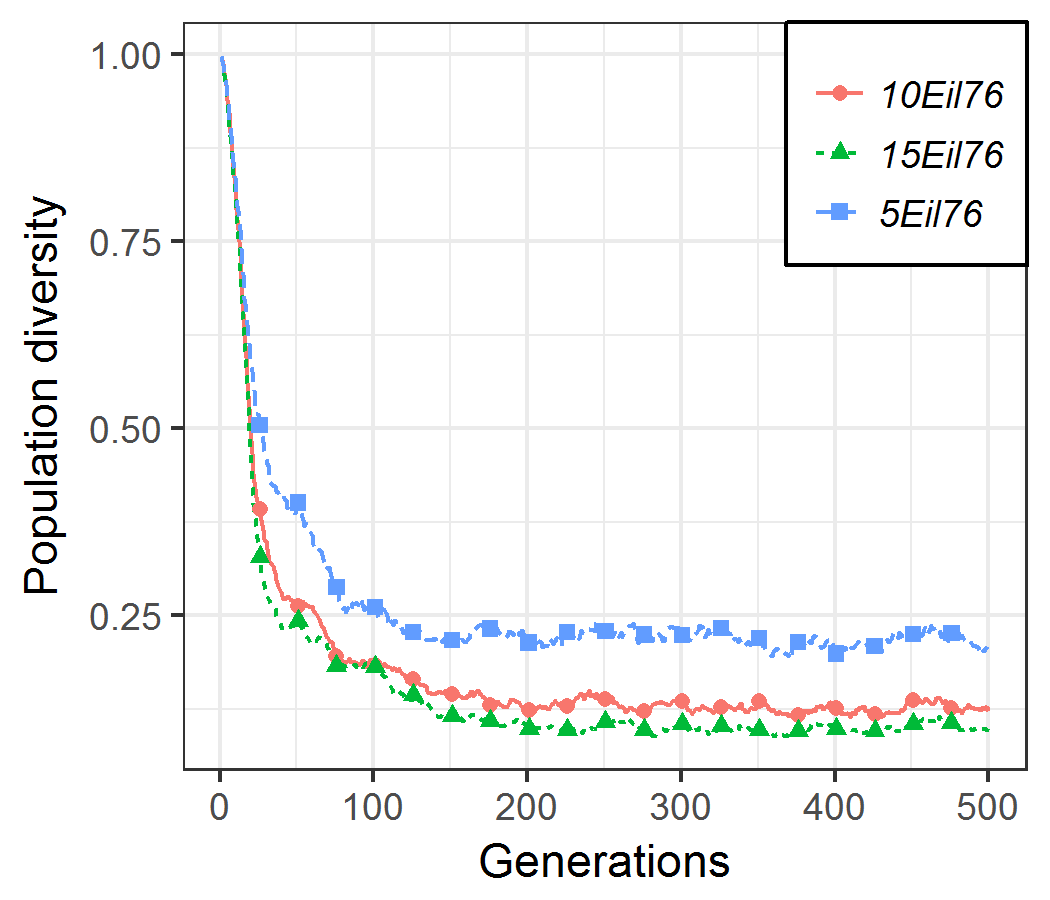}
		\caption{Type 1}
		\label{fig:PopulationDiversity-Type1}
	\end{subfigure}
	\begin{subfigure}[b]{.48\linewidth}
		\centering
		\renewcommand{\scalefigure}{0.55}
		\includegraphics[scale=\scalefigure]{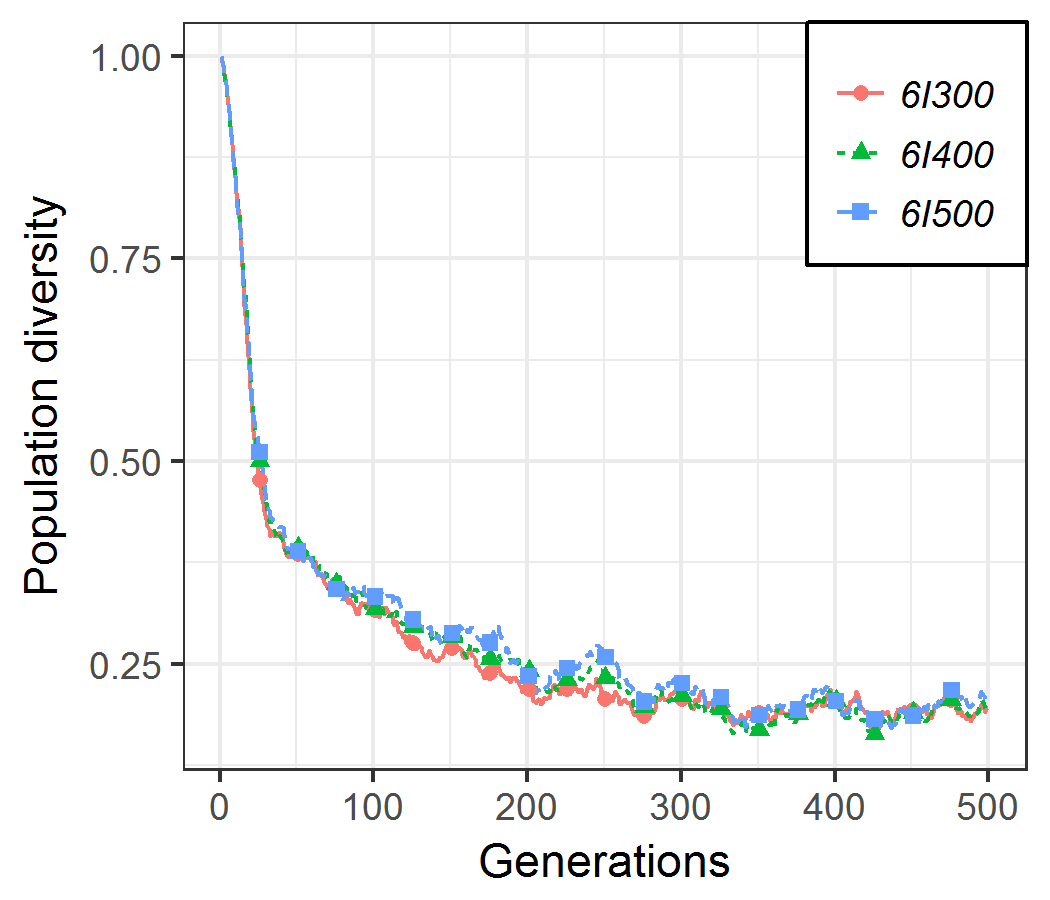}
		\caption{Type 3}
		\label{figPopulationDiversity-Type3}
	\end{subfigure}
	\begin{subfigure}[b]{.48\linewidth}
		\centering
		\renewcommand{\scalefigure}{0.55}
		\includegraphics[scale=\scalefigure]{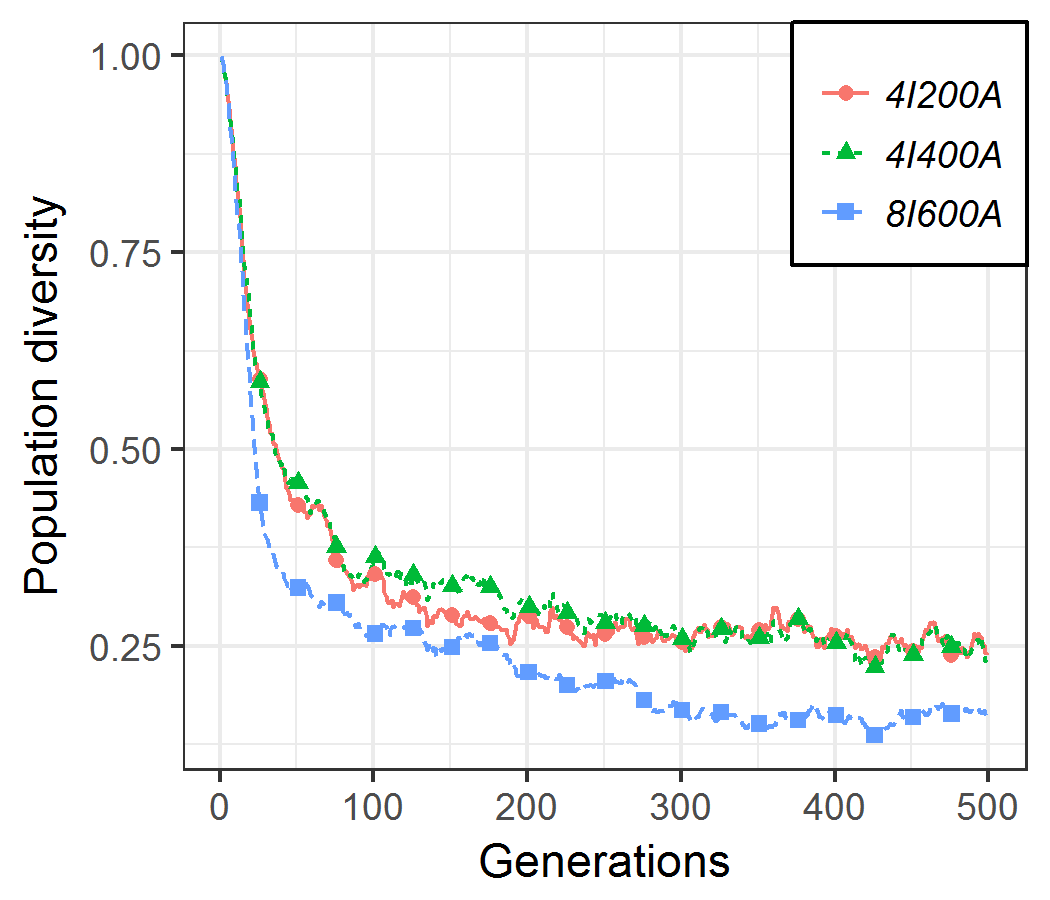}
		\caption{Type 4}
		\label{figPopulationDiversity-Type4}
	\end{subfigure}
	\begin{subfigure}[b]{.48\linewidth}
		\centering
		\renewcommand{\scalefigure}{0.55}
		\includegraphics[scale=\scalefigure]{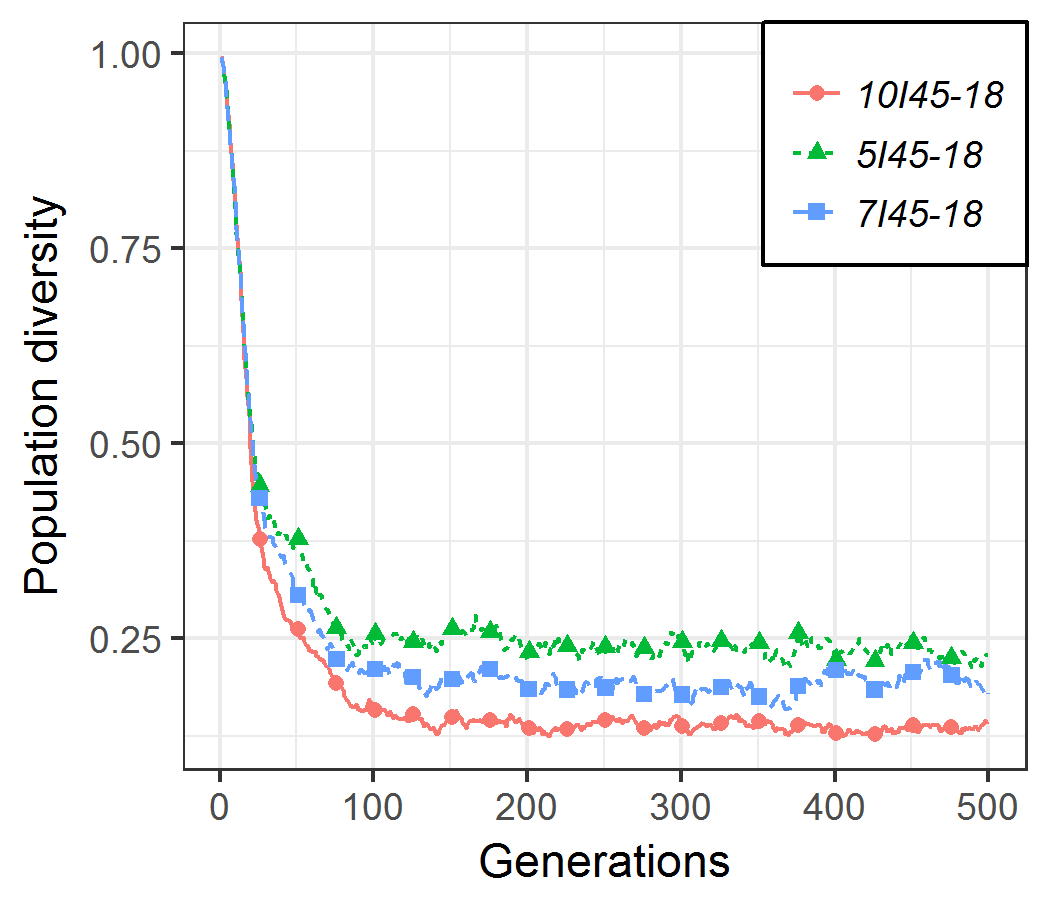}
		\caption{Type 5}
		\label{fig:PopulationDiversity-Type5}
	\end{subfigure}
	\begin{subfigure}[b]{.48\linewidth}
		\centering
		\renewcommand{\scalefigure}{0.55}
		\includegraphics[scale=\scalefigure]{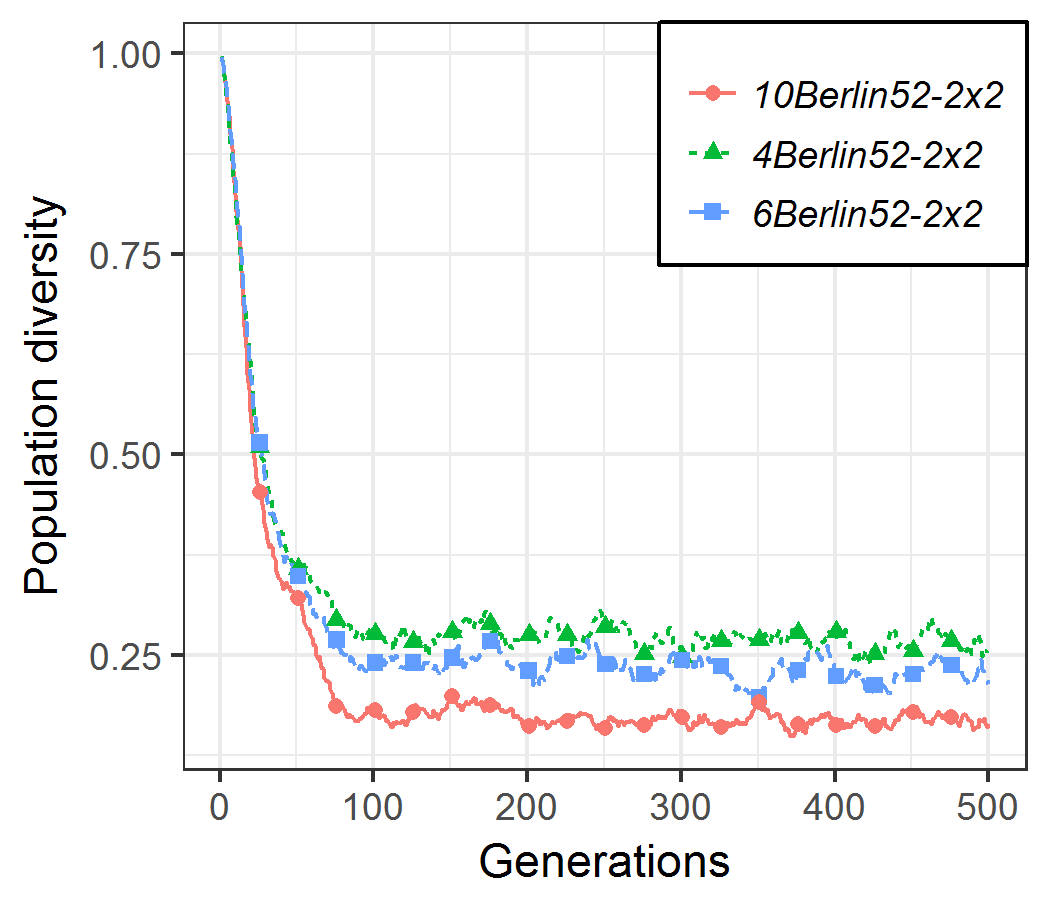}
		\caption{Type 6}
		\label{fig:PopulationDiversity-Type6}
	\end{subfigure}
	\caption{Relationship between polulation diversity and generation in types}
	\label{fig:Relationship-Bbetween-Polulation-Diversity-and-Generation}
\end{figure}
\setlength{\intextsep}{0pt}

Figure~\ref{fig:Relationship-Bbetween-Polulation-Diversity-and-Generation} illustrates the relationship between the population diversity and the number of generations on instances in types. A highlight in these figures is that with same number of vertices, the larger the population diversity at a generation, the smaller the number of clusters. For example, in Figure~\ref{fig:PopulationDiversity-Type5}, the number of clusters in instance 5i45-18 is the smallest (5 clusters), while the corresponding population diversity is the largest. While the number of clusters in instance 10i45-18 is the largest (10 clusters), the corresponding population diversity is the smallest.

\section{Conclusion}
\label{Sec_Conclusion}
This paper presented a new approach that decompose the \gls{clustp} into two sub-problems. This paper also pointed out two features of these sub-problems which help to find a solution of the \gls{clustp} by using the properties of solutions to the sub-problems, i.e. set of edges connecting among clusters and set of roots of clusters. From these observations, we proposed an evolutionary algorithm based on a new individual encoding. Various types of problem instances are selected to conduct the evolutionary algorithm. The experimental results demonstrated superior performance of the new algorithm in comparison with existing algorithms. An in-depth analysis of the received results also explained the impact of the average number of vertices in a cluster and the number of clusters to the efficiency of the new algorithm. More specifically, the average number of vertices in a cluster is has more impact than the number of clusters.

To enhance the performance of the novel algorithm, in the future, we will find a local search algorithm to combine with this evolutionary algorithm.



\clearpage
\onecolumn
\begin{landscape}
	
\begin{ThreePartTable}
	\begin{TableNotes}
		\footnotesize
		\item Rm: Running time of algorithms in minutes; \hspace{1cm}  Ins: Problem Instances; 
	\end{TableNotes}
	%
	\begin{longtable}{|c|l|r|r|c|r|r|c|r|r|r|}
		\caption{Results Obtained By E-MFEA, C-MFEA And NEA on Instances In Type 1} \label{tab:ResultsType1}\\
    \hline
    \multirow{2}[4]{*}{} & 
    \multicolumn{1}{c|}{\multirow{2}[4]{*}{\textbf{Ins}}} & 
    \multicolumn{3}{c|}{\textbf{E-MFEA}} & 
    \multicolumn{3}{c|}{\textbf{C-MFEA}} & 
    \multicolumn{3}{c|}{\textbf{NEA}} \\
    
    \cline{3-11}          &       & 
    \multicolumn{1}{c|}{\textbf{BF}} & 
    \multicolumn{1}{c|}{\textbf{Avg}} & 
    \multicolumn{1}{p{1.3em}|}{\textbf{Rm}}& 
    \multicolumn{1}{c|}{\textbf{BF}} & 
    \multicolumn{1}{c|}{\textbf{Avg}} & 
    \multicolumn{1}{p{1.3em}|}{\textbf{Rm}} & 
    \multicolumn{1}{c|}{\textbf{BF}} & 
    \multicolumn{1}{c|}{\textbf{Avg}} & 
    \multicolumn{1}{p{1.3em}|}{\textbf{Rm}}\\
    
    \hline
    \multirow{27}[54]{*}{\begin{sideways}\textbf{Small Instances}\end{sideways}} & 10berlin52 & 45684.5 & 47075.7 & 0.10  & -     & -     & -     & 43954.0 & 44237.6 & 0.02 \\
\cline{2-11}          & 10eil51 & 1923.9 & 2020.3 & 0.10  & 3027.7 & 3513.3 & 0.05  & 1741.5 & 1770.6 & 0.02 \\
\cline{2-11}          & 10eil76 & 3089.3 & 3418.7 & 0.22  & 4263.1 & 5175.2 & 0.05  & 2264.5 & 2315.6 & 0.02 \\
\cline{2-11}          & 10kroB100 & 203481.4 & 221058.2 & 0.22  & 301982.8 & 363749.0 & 0.08  & 143108.6 & 147539.7 & 0.02 \\
\cline{2-11}          & 10pr76 & 656800.9 & 685310.3 & 0.22  & 914315.2 & 1159871.7 & 0.08  & 531536.7 & 544954.5 & 0.02 \\
\cline{2-11}          & 10rat99 & 10381.7 & 11015.9 & 0.22  & 13954.6 & 17406.1 & 0.07  & 7697.8 & 7899.4 & 0.02 \\
\cline{2-11}          & 10st70 & 5723.3 & 5833.4 & 0.17  & -     & -     & -     & 3098.7 & 3191.1 & 0.02 \\
\cline{2-11}          & 15berlin52 & 29246.2 & 30279.9 & 0.17  & -     & -     & -     & 26463.1 & 26867.8 & 0.03 \\
\cline{2-11}          & 15eil51 & 1753.5 & 1954.1 & 0.15  & -     & -     & -     & 1313.4 & 1336.5 & 0.03 \\
\cline{2-11}          & 15eil76 & 3374.9 & 3452.4 & 0.15  & -     & -     & -     & 2955.3 & 3047.8 & 0.03 \\
\cline{2-11}          & 15pr76 & 772012.8 & 796271.2 & 0.15  & -     & -     & -     & 714652.2 & 728128.0 & 0.03 \\
\cline{2-11}          & 15st70 & 4921.2 & 5308.3 & 0.15  & -     & -     & -     & 4145.8 & 4230.1 & 0.03 \\
\cline{2-11}          & 25eil101 & 5241.4 & 5384.7 & 0.27  & -     & -     & -     & 4826.6 & 4885.5 & 0.03 \\
\cline{2-11}          & 25kroA100 & 165880.9 & 169702.2 & 0.27  & -     & -     & -     & 150157.7 & 153155.6 & 0.03 \\
\cline{2-11}          & 25lin105 & 107677.9 & 110598.5 & 0.30  & -     & -     & -     & 98991.8 & 100615.8 & 0.03 \\
\cline{2-11}          & 25rat99 & 9464.2 & 9690.5 & 0.30  & -     & -     & -     & 7056.0 & 7162.3 & 0.03 \\
\cline{2-11}          & 50eil101 & 4239.2 & 4459.4 & 0.37  & -     & -     & -     & 3890.7 & 3919.7 & 0.07 \\
\cline{2-11}          & 50kroA100 & 180990.7 & 199637.3 & 0.37  & -     & -     & -     & 160547.4 & 161889.6 & 0.07 \\
\cline{2-11}          & 50kroB100 & 156209.3 & 170468.1 & 0.37  & -     & -     & -     & 134077.5 & 135332.2 & 0.07 \\
\cline{2-11}          & 50lin105 & 153465.7 & 158775.5 & 0.37  & -     & -     & -     & 146367.1 & 147175.4 & 0.07 \\
\cline{2-11}          & 50rat99 & 9747.3 & 11328.3 & 0.82  & -     & -     & -     & 8104.5 & 8132.4 & 0.08 \\
\cline{2-11}          & 5berlin52 & 35387.5 & 37595.9 & 0.82  & 42296.7 & 48591.5 & 0.07  & 22746.4 & 22938.2 & 0.00 \\
\cline{2-11}          & 5eil51 & 2101.3 & 2367.0 & 0.18  & 2380.1 & 2691.7 & 0.05  & 1769.4 & 1775.3 & 0.00 \\
\cline{2-11}          & 5eil76 & 3450.1 & 3688.0 & 0.18  & 4962.0 & 5583.6 & 0.05  & 2630.8 & 2693.1 & 0.00 \\
\cline{2-11}          & 5pr76 & 709511.2 & 799642.4 & 0.23  & 1056191.9 & 1261431.3 & 0.05  & 585008.0 & 591547.0 & 0.00 \\
\cline{2-11}          & 5st70 & 5430.2 & 5693.8 & 0.23  & 6598.6 & 7550.2 & 0.05  & 4520.1 & 4544.9 & 0.00 \\
    \hline
    \multirow{15}[30]{*}{\begin{sideways}\textbf{Large Instances}\end{sideways}} & 10a280 & 60569.9 & 65704.5 & 0.13  & 101817.0 & 118642.0 & 0.03  & 28690.9 & 29664.8 & 0.02 \\
\cline{2-11}          & 10gil262 & 49367.9 & 52619.9 & 0.13  & -     & -     & -     & 29075.0 & 29568.4 & 0.02 \\
\cline{2-11}          & 10lin318 & 1271784.7 & 1360198.6 & 0.28  & 2053763.5 & 2289808.8 & 0.03  & 832299.5 & 841893.2 & 0.02 \\
\cline{2-11}          & 10pcb442 & 1478189.3 & 1627962.0 & 0.28  & 3037722.6 & 3339950.7 & 0.07  & 765561.0 & 796960.4 & 0.02 \\
\cline{2-11}          & 10pr439 & 2715809.1 & 2927551.1 & 0.32  & 7422453.2 & 8398138.5 & 0.07  & 1971633.0 & 2022257.4 & 0.02 \\
\cline{2-11}          & 25a280 & 46362.3 & 50294.1 & 0.32  & 134286.1 & 166395.9 & 0.03  & 31481.2 & 32020.2 & 0.03 \\
\cline{2-11}          & 25gil262 & 40597.5 & 42488.2 & 0.08  & 116170.2 & 136005.8 & 0.03  & 31579.5 & 31949.7 & 0.03 \\
\cline{2-11}          & 25lin318 & 848201.5 & 927441.2 & 0.08  & 2449806.0 & 2883914.2 & 0.05  & 607029.0 & 617399.9 & 0.03 \\
\cline{2-11}          & 25pcb442 & 1171370.9 & 1253188.9 & 0.42  & 3346967.2 & 4070674.9 & 0.05  & 794217.4 & 805896.7 & 0.03 \\
\cline{2-11}          & 25pr439 & 2206210.6 & 2399266.5 & 0.42  & -     & -     & -     & 1585283.0 & 1612334.7 & 0.03 \\
\cline{2-11}          & 50a280 & 48703.5 & 52488.9 & 0.12  & -     & -     & -     & 37458.4 & 37828.6 & 0.10 \\
\cline{2-11}          & 50gil262 & 34324.7 & 35734.5 & 0.12  & -     & -     & -     & 27647.5 & 27836.2 & 0.10 \\
\cline{2-11}          & 50lin318 & 825219.7 & 881360.3 & 0.23  & -     & -     & -     & 706854.9 & 713744.5 & 0.10 \\
\cline{2-11}          & 50pcb442 & 1194878.5 & 1313254.1 & 0.23  & -     & -     & -     & 949830.8 & 954169.0 & 0.10 \\
    \hline
    \insertTableNotes	  
\end{longtable}%
\end{ThreePartTable}	
	
	\begin{ThreePartTable}
	\begin{TableNotes}
		\footnotesize
		\item Rm: Running time of algorithms in minutes; \hspace{1cm}  Ins: Problem Instances; 
	\end{TableNotes}
 \begin{longtable}{|c|l|r|r|c|r|r|c|r|r|r|}
	\caption{Results Obtained By E-MFEA, C-MFEA And NEA on Instances In Type 3 and Type 4} \label{tab:ResultsType34}\\
	\hline
	\multirow{2}[4]{*}{} & 
	\multicolumn{1}{c|}{\multirow{2}[4]{*}{\textbf{Ins}}} & 
	\multicolumn{3}{c|}{\textbf{E-MFEA}} & 
	\multicolumn{3}{c|}{\textbf{C-MFEA}} & 
	\multicolumn{3}{c|}{\textbf{NEA}} \\
	
	\cline{3-11}          &       & 
	\multicolumn{1}{c|}{\textbf{BF}} & 
	\multicolumn{1}{c|}{\textbf{Avg}} & 
	\multicolumn{1}{p{1.3em}|}{\textbf{Rm}}& 
	\multicolumn{1}{c|}{\textbf{BF}} & 
	\multicolumn{1}{c|}{\textbf{Avg}} & 
	\multicolumn{1}{p{1.3em}|}{\textbf{Rm}} & 
	\multicolumn{1}{c|}{\textbf{BF}} & 
	\multicolumn{1}{c|}{\textbf{Avg}} & 
	\multicolumn{1}{p{1.3em}|}{\textbf{Rm}}\\

    \hline
    \multirow{5}[10]{*}{\begin{sideways}\textbf{Type 3}\end{sideways}} & 6i300 & \multicolumn{1}{c|}{28672.7} & \multicolumn{1}{c|}{30356.7} & 7.38  & \multicolumn{1}{c|}{31283.6} & \multicolumn{1}{c|}{39444.1} & 1.22  & 19358.8 & 19467.0 & 0.02 \\
		\cline{2-11}          & 6i350 & \multicolumn{1}{c|}{32978.7} & \multicolumn{1}{c|}{34541.8} & 7.38  & \multicolumn{1}{c|}{78019.0} & \multicolumn{1}{c|}{87416.8} & 1.22  & 21472.8 & 21702.2 & 0.02 \\
		\cline{2-11}          & 6i400 & \multicolumn{1}{c|}{43086.1} & \multicolumn{1}{c|}{46328.2} & 10.47 & \multicolumn{1}{c|}{52463.2} & \multicolumn{1}{c|}{62720.6} & 2.05  & 29506.9 & 29677.7 & 0.02 \\
		\cline{2-11}          & 6i450 & \multicolumn{1}{c|}{51738.6} & \multicolumn{1}{c|}{55060.5} & 10.47 & \multicolumn{1}{c|}{124525.9} & \multicolumn{1}{c|}{137183.6} & 2.05  & 35866.3 & 36124.5 & 0.02 \\
		\cline{2-11}          & 6i500 & \multicolumn{1}{c|}{61227.4} & \multicolumn{1}{c|}{69324.1} & 5.88  & \multicolumn{1}{c|}{143220.7} & \multicolumn{1}{c|}{153699.3} & 2.93  & 37711.6 & 38045.9 & 0.02 \\
		    \hline
		    \multirow{10}[20]{*}{\begin{sideways}\textbf{Type 4}\end{sideways}} & 4i200a & 329844.5 & 393407.0 & 0.07  & 1058533.6 & 1150968.4 & 0.02  & 97959.6 & 102256.3 & 0.00 \\
		\cline{2-11}          & 4i200h & 215812.7 & 248321.3 & 0.07  & 604259.8 & 652916.9 & 0.02  & 87675.3 & 89628.9 & 0.00 \\
		\cline{2-11}          & 4i200x1 & 239439.4 & 258423.4 & 0.07  & 621807.6 & 680006.8 & 0.02  & 123669.7 & 125782.7 & 0.00 \\
		\cline{2-11}          & 4i200x2 & 226820.7 & 254560.0 & 0.07  & 631471.0 & 671268.7 & 0.02  & 114012.3 & 116256.5 & 0.00 \\
		\cline{2-11}          & 4i200z & 259796.7 & 284456.1 & 0.40  & 265010.4 & 284466.7 & 0.05  & 131683.5 & 133873.8 & 0.00 \\
		\cline{2-11}          & 4i400a & 834356.4 & 948781.5 & 0.40  & 4681600.4 & 5013170.5 & 0.05  & 217171.4 & 227530.6 & 0.02 \\
		\cline{2-11}          & 4i400h & 655591.7 & 713631.1 & 5.15  & 2602030.7 & 2808230.9 & 3.05  & 257954.4 & 260916.0 & 0.02 \\
		\cline{2-11}          & 4i400x1 & 569429.8 & 667408.9 & 5.15  & 2481894.5 & 2668551.2 & 3.05  & 188786.2 & 191694.7 & 0.02 \\
		\cline{2-11}          & 4i400x2 & 570171.7 & 639480.4 & 5.50  & 2478070.1 & 2659442.4 & 3.13  & 159254.8 & 163222.8 & 0.02 \\
		\cline{2-11}          & 4i400z & 588799.3 & 674446.0 & 5.50  & 2512497.8 & 2723834.3 & 3.13  & 221460.6 & 225096.3 & 0.02 \\
    	\hline
	  \insertTableNotes	  
	\end{longtable}%
\end{ThreePartTable}	
	
\begin{ThreePartTable}
	\begin{TableNotes}
		\footnotesize
		\item Rm: Running time of algorithms in minutes; \hspace{1cm}  Ins: Problem Instances; 
	\end{TableNotes}
%
    \begin{longtable}{|c|l|r|r|c|r|r|c|r|r|r|}
    	\caption{Results Obtained By E-MFEA, C-MFEA And NEA on Instances In Type 5} \label{tab:ResultsType5}\\
	    \hline
	    \multirow{2}[4]{*}{} & 
	    \multicolumn{1}{c|}{\multirow{2}[4]{*}{\textbf{Ins}}} & 
	    \multicolumn{3}{c|}{\textbf{E-MFEA}} & 
	    \multicolumn{3}{c|}{\textbf{C-MFEA}} & 
	    \multicolumn{3}{c|}{\textbf{NEA}} \\
	    
		\cline{3-11}          &       & 
		\multicolumn{1}{c|}{\textbf{BF}} & 
		\multicolumn{1}{c|}{\textbf{Avg}} & 
		\multicolumn{1}{p{1.3em}|}{\textbf{Rm}}& 
		\multicolumn{1}{c|}{\textbf{BF}} & 
		\multicolumn{1}{c|}{\textbf{Avg}} & 
		\multicolumn{1}{p{1.3em}|}{\textbf{Rm}} & 
		\multicolumn{1}{c|}{\textbf{BF}} & 
		\multicolumn{1}{c|}{\textbf{Avg}} & 
		\multicolumn{1}{p{1.3em}|}{\textbf{Rm}}\\
		
	    \hline
	    \multirow{21}[42]{*}{\begin{sideways}\textbf{Small Instances}\end{sideways}} & 
	   		10i120-46 & 122349.2 & 125510.1 & 0.32  & 156097.1 & 184275.7 & 0.10  & 96168.2 & 97752.1 & 0.02 \\
			\cline{2-11}          & 10i30-17 & 14718.5 & 15740.9 & 0.32  & -     & -     & -     & 13276.6 & 13290.0 & 0.02 \\
			\cline{2-11}          & 10i45-18 & 27068.0 & 29306.5 & 0.12  & 37121.1 & 42932.6 & 0.10  & 23227.3 & 23985.5 & 0.02 \\
			\cline{2-11}          & 10i60-21 & 42386.6 & 44667.3 & 0.12  & 53877.1 & 68825.7 & 0.05  & 34147.0 & 35233.3 & 0.02 \\
			\cline{2-11}          & 10i65-21 & 47165.1 & 50815.3 & 0.15  & 66268.0 & 79374.5 & 0.05  & 38318.8 & 39578.8 & 0.02 \\
			\cline{2-11}          & 10i70-21 & 48058.1 & 51889.0 & 0.15  & 61048.5 & 76907.1 & 0.08  & 38816.6 & 39687.3 & 0.02 \\
			\cline{2-11}          & 10i75-22 & 74952.5 & 77605.4 & 0.25  & -     & -     & -     & 65923.2 & 66485.1 & 0.02 \\
			\cline{2-11}          & 10i90-33 & 66438.4 & 67881.2 & 0.25  & 81379.3 & 97534.6 & 0.08  & 53076.0 & 54636.2 & 0.02 \\
			\cline{2-11}          & 5i120-46 & 92826.2 & 103713.6 & 0.47  & -     & -     & -     & 61695.7 & 62620.1 & 0.02 \\
			\cline{2-11}          & 5i30-17 & 15801.5 & 17664.7 & 0.47  & -     & -     & -     & 14399.9 & 14399.9 & 0.00 \\
			\cline{2-11}          & 5i45-18 & 19813.0 & 23639.3 & 0.12  & 24131.3 & 27649.3 & 0.05  & 14884.3 & 14925.6 & 0.00 \\
			\cline{2-11}          & 5i60-21 & 36445.8 & 39060.5 & 0.12  & 48867.1 & 57579.8 & 0.05  & 28422.7 & 28769.6 & 0.00 \\
			\cline{2-11}          & 5i65-21 & 38682.7 & 41488.6 & 0.15  & 51818.2 & 62189.7 & 0.07  & 30907.8 & 31254.4 & 0.00 \\
			\cline{2-11}          & 5i70-21 & 50025.1 & 54839.7 & 0.15  & 69390.4 & 82771.9 & 0.07  & 35052.8 & 35298.8 & 0.00 \\
			\cline{2-11}          & 5i75-22 & 41260.0 & 49758.6 & 0.27  & 66131.0 & 78693.8 & 0.08  & 34692.5 & 35098.7 & 0.00 \\
			\cline{2-11}          & 5i90-33 & 69640.4 & 74725.8 & 0.27  & 91746.3 & 99266.9 & 0.08  & 51977.0 & 52533.8 & 0.00 \\
			\cline{2-11}          & 7i30-17 & 24546.7 & 26344.9 & 0.10  & -     & -     & -     & 20438.9 & 20454.2 & 0.02 \\
			\cline{2-11}          & 7i45-18 & 32673.8 & 34086.1 & 0.10  & -     & -     & -     & 20512.0 & 20700.8 & 0.02 \\
			\cline{2-11}          & 7i60-21 & 45073.1 & 48395.9 & 0.15  & 55312.7 & 67556.5 & 0.05  & 36295.4 & 37780.7 & 0.02 \\
			\cline{2-11}          & 7i65-21 & 47276.4 & 49872.8 & 0.15  & 58179.2 & 71715.7 & 0.05  & 35201.2 & 36136.4 & 0.02 \\
			\cline{2-11}          & 7i70-21 & 54019.4 & 60450.6 & 13.47 & 57464.9 & 67043.4 & 1.10  & 39613.4 & 40819.5 & 0.02 \\
			    \hline
			    \multirow{15}[30]{*}{\begin{sideways}\textbf{Large Instances}\end{sideways}} & 10i300-109 & 202810.9 & 217296.6 & 0.22  & 255346.6 & 312345.5 & 0.07  & 117421.2 & 119952.9 & 0.02 \\
			\cline{2-11}          & 10i400-206 & 309133.7 & 335121.6 & 0.22  & 591868.0 & 695950.3 & 0.07  & 214604.4 & 217399.0 & 0.02 \\
			\cline{2-11}          & 10i500-305 & 455477.7 & 476345.7 & 0.32  & 835260.4 & 921942.4 & 0.10  & 355952.4 & 359614.5 & 0.02 \\
			\cline{2-11}          & 15i300-110 & 198774.7 & 216560.0 & 0.32  & 341726.8 & 403762.5 & 0.10  & 119922.2 & 122146.2 & 0.02 \\
			\cline{2-11}          & 15i400-207 & 277578.7 & 301202.3 & 0.22  & 523177.2 & 635130.8 & 0.08  & 171349.6 & 175081.7 & 0.02 \\
			\cline{2-11}          & 15i500-306 & 461867.3 & 490046.0 & 0.22  & 930442.1 & 1142750.5 & 0.00  & 310122.7 & 313184.9 & 0.02 \\
			\cline{2-11}          & 20i300-111 & 227654.2 & 243449.4 & 0.13  & 498896.3 & 648713.3 & 0.05  & 163927.8 & 167104.4 & 0.03 \\
			\cline{2-11}          & 20i400-208 & 315302.4 & 328243.0 & 3.78  & 798929.3 & 996341.1 & 1.60  & 231753.3 & 236373.9 & 0.03 \\
			\cline{2-11}          & 20i500-307 & 325116.8 & 361197.5 & 6.38  & 546303.4 & 631936.3 & 2.20  & 212306.3 & 215644.7 & 0.03 \\
			\cline{2-11}          & 25i300-112 & 198094.7 & 212148.3 & 6.38  & 538605.8 & 636235.0 & 2.20  & 125392.7 & 127466.5 & 0.05 \\
			\cline{2-11}          & 25i400-209 & 358815.6 & 385364.4 & 5.87  & 874763.8 & 1109608.0 & 2.27  & 241529.0 & 243994.1 & 0.05 \\
			\cline{2-11}          & 25i500-308 & 413415.1 & 430338.8 & 5.87  & 1154806.9 & 1368867.4 & 2.27  & 312805.6 & 316116.8 & 0.05 \\
			\cline{2-11}          & 5i300-108 & 276565.1 & 288332.1 & 6.72  & 267454.9 & 298521.2 & 1.53  & 178628.1 & 180397.2 & 0.02 \\
			\cline{2-11}          & 5i400-205 & 321381.9 & 346678.0 & 6.72  & 841789.9 & 890827.0 & 1.53  & 211603.0 & 213125.6 & 0.02 \\
			\cline{2-11}          & 5i500-304 & 321328.5 & 345679.7 & 13.67 & 959132.5 & 1050820.4 & 2.93  & 183656.4 & 185924.4 & 0.02 \\
	    \hline
	  \insertTableNotes	  
	\end{longtable}%
\end{ThreePartTable}

\begin{ThreePartTable}
	\begin{TableNotes}
		\footnotesize
		\item Rm: Running time of algorithms in minutes; \hspace{1cm}  Ins: Problem Instances; 
	\end{TableNotes}
	%
	\begin{longtable}{|c|l|r|r|c|r|r|c|r|r|r|}
		\caption{Results Obtained By E-MFEA, C-MFEA And NEA on Instances In Type 1} \label{tab:ResultsType6}\\
		\hline
		\multirow{2}[4]{*}{} & 
		\multicolumn{1}{c|}{\multirow{2}[4]{*}{\textbf{Ins}}} & 
		\multicolumn{3}{c|}{\textbf{E-MFEA}} & 
		\multicolumn{3}{c|}{\textbf{C-MFEA}} & 
		\multicolumn{3}{c|}{\textbf{NEA}} \\
		
		\cline{3-11}          &       & 
		\multicolumn{1}{c|}{\textbf{BF}} & 
		\multicolumn{1}{c|}{\textbf{Avg}} & 
		\multicolumn{1}{p{1.3em}|}{\textbf{Rm}}& 
		\multicolumn{1}{c|}{\textbf{BF}} & 
		\multicolumn{1}{c|}{\textbf{Avg}} & 
		\multicolumn{1}{p{1.3em}|}{\textbf{Rm}} & 
		\multicolumn{1}{c|}{\textbf{BF}} & 
		\multicolumn{1}{c|}{\textbf{Avg}} & 
		\multicolumn{1}{p{1.3em}|}{\textbf{Rm}}\\
		
	    \hline
	    \multirow{36}[72]{*}{\begin{sideways}\textbf{Small Instances}\end{sideways}} & 10berlin52-2x5 & 34749.2 & 36828.8 & 1.60  & -     & -     & -     & 27471.4 & 27805.3 & 0.02 \\
		\cline{2-11}          & 12eil51-3x4 & 1922.4 & 2000.3 & 0.18  & 3115.8 & 3648.4 & 0.07  & 1720.1 & 1762.7 & 0.02 \\
		\cline{2-11}          & 12eil76-3x4 & 3197.7 & 3330.0 & 0.18  & 5219.5 & 6381.5 & 0.07  & 2738.6 & 2802.0 & 0.02 \\
		\cline{2-11}          & 12pr76-3x4 & 699229.9 & 723373.3 & 0.22  & -     & -     & -     & 604837.0 & 621228.6 & 0.02 \\
		\cline{2-11}          & 12st70-3x4 & 5113.1 & 5431.3 & 0.22  & 6976.2 & 8579.5 & 0.12  & 4148.4 & 4219.2 & 0.02 \\
		\cline{2-11}          & 15pr76-3x5 & 560767.5 & 576792.5 & 0.23  & -     & -     & -     & 534613.0 & 544174.0 & 0.03 \\
		\cline{2-11}          & 16eil51-4x4 & 1459.1 & 1490.5 & 0.23  & -     & -     & -     & 1323.8 & 1351.1 & 0.03 \\
		\cline{2-11}          & 16eil76-4x4 & 3321.5 & 3453.1 & 0.38  & -     & -     & -     & 2088.1 & 2163.0 & 0.03 \\
		\cline{2-11}          & 16lin105-4x4 & 158387.1 & 162660.9 & 0.38  & -     & -     & -     & 128713.1 & 130815.3 & 0.03 \\
		\cline{2-11}          & 16st70-4x4 & 3410.2 & 3519.6 & 0.25  & -     & -     & -     & 2963.8 & 3050.4 & 0.03 \\
		\cline{2-11}          & 18pr76-3x6 & 735572.7 & 771984.5 & 0.25  & -     & -     & -     & 641209.6 & 657524.3 & 0.03 \\
		\cline{2-11}          & 20eil51-4x5 & 2583.1 & 2621.0 & 0.22  & -     & -     & -     & 2286.4 & 2331.3 & 0.03 \\
		\cline{2-11}          & 20eil76-4x5 & 2886.0 & 3038.8 & 0.22  & -     & -     & -     & 2478.2 & 2520.3 & 0.02 \\
		\cline{2-11}          & 20st70-4x5 & 4387.9 & 4582.7 & 0.38  & -     & -     & -     & 2976.9 & 3032.7 & 0.03 \\
		\cline{2-11}          & 25eil101-5x5 & 4314.2 & 4546.4 & 0.38  & -     & -     & -     & 3711.7 & 3780.9 & 0.03 \\
		\cline{2-11}          & 25eil51-5x5 & 1550.4 & 1640.2 & 0.22  & -     & -     & -     & 1483.7 & 1512.0 & 0.03 \\
		\cline{2-11}          & 25eil76-5x5 & 3523.0 & 3722.1 & 0.22  & -     & -     & -     & 2264.3 & 2312.1 & 0.03 \\
		\cline{2-11}          & 25rat99-5x5 & 12283.6 & 12635.4 & 0.45  & -     & -     & -     & 11754.1 & 11869.1 & 0.03 \\
		\cline{2-11}          & 28kroA100-4x7 & 161087.0 & 173691.0 & 0.45  & -     & -     & -     & 138682.6 & 141334.2 & 0.03 \\
		\cline{2-11}          & 2lin105-2x1 & 300290.7 & 319749.4 & 1.60  & 920629.6 & 1035690.4 & 0.12  & 152729.7 & 152729.7 & 0.00 \\
		\cline{2-11}          & 30kroB100-5x6 & 216499.5 & 227926.8 & 0.62  & -     & -     & -     & 201813.7 & 204967.3 & 0.03 \\
		\cline{2-11}          & 35kroB100-5x5 & 166362.5 & 179525.2 & 0.92  & -     & -     & -     & 133662.2 & 137003.4 & 0.03 \\
		\cline{2-11}          & 36eil101-6x6 & 4752.6 & 5226.1 & 0.62  & -     & -     & -     & 3977.6 & 4028.6 & 0.05 \\
		\cline{2-11}          & 42rat99-6x7 & 9706.4 & 10068.5 & 0.92  & -     & -     & -     & 9093.5 & 9182.7 & 0.07 \\
		\cline{2-11}          & 4berlin52-2x2 & 37576.7 & 43289.4 & 0.25  & 58055.8 & 65276.7 & 0.05  & 23287.9 & 23287.9 & 0.00 \\
		\cline{2-11}          & 4eil51-2x2 & 2691.4 & 2870.8 & 0.25  & 2866.8 & 3200.3 & 0.05  & 1898.5 & 1901.3 & 0.00 \\
		\cline{2-11}          & 4eil76-2x2 & 4312.7 & 4800.2 & 0.53  & 5736.3 & 6451.4 & 0.08  & 2948.7 & 2955.8 & 0.00 \\
		\cline{2-11}          & 4pr76-2x2 & 747062.3 & 821661.9 & 0.53  & 1227486.9 & 1437245.4 & 0.08  & 442693.0 & 446682.1 & 0.00 \\
		\cline{2-11}          & 6berlin52-2x3 & 40772.6 & 43360.3 & 0.38  & -     & -     & -     & 32128.6 & 32354.7 & 0.00 \\
		\cline{2-11}          & 6pr76-2x3 & 747967.5 & 822531.1 & 0.38  & 1049553.9 & 1226278.5 & 0.07  & 648713.1 & 659658.8 & 0.00 \\
		\cline{2-11}          & 6st70-2x3 & 4287.0 & 4631.6 & 0.32  & 5354.6 & 6458.5 & 0.07  & 3478.2 & 3525.3 & 0.00 \\
		\cline{2-11}          & 8berlin52-2x4 & 35300.7 & 40698.9 & 0.32  & -     & -     & -     & 26783.2 & 27060.4 & 0.02 \\
		\cline{2-11}          & 9eil101-3x3 & 4397.4 & 4893.6 & 0.38  & 6491.1 & 8198.1 & 0.08  & 3184.4 & 3274.5 & 0.02 \\
		\cline{2-11}          & 9eil51-3x3 & 2197.0 & 2293.3 & 0.38  & 2841.6 & 3382.1 & 0.08  & 1916.0 & 1963.8 & 0.02 \\
		\cline{2-11}          & 9eil76-3x3 & 3761.3 & 3880.8 & 0.32  & 4945.0 & 5768.9 & 0.08  & 2999.4 & 3057.2 & 0.02 \\
		\cline{2-11}          & 9pr76-3x3 & 738481.6 & 778897.8 & 0.32  & 1012732.3 & 1227532.7 & 0.08  & 558349.3 & 567987.4 & 0.02 \\
		    \hline
		    \multirow{14}[28]{*}{\begin{sideways}\textbf{Large Instances}\end{sideways}} & 18pr439-3x6 & 2965170.6 & 3424145.3 & 23.60 & -     & -     & -     & 1525370.2 & 1553940.8 & 0.03 \\
		\cline{2-11}          & 20pr439-4x5 & 3280842.4 & 3597063.5 & 23.60 & -     & -     & -     & 2035939.4 & 2075349.7 & 0.05 \\
		\cline{2-11}          & 25a280-5x5 & 54738.3 & 59798.3 & 2.03  & 142031.1 & 166462.8 & 0.93  & 43408.0 & 44268.6 & 0.05 \\
		\cline{2-11}          & 25gil262-5x5 & 41947.3 & 44751.4 & 2.03  & 104594.3 & 128710.8 & 0.93  & 32172.6 & 32674.3 & 0.05 \\
		\cline{2-11}          & 25pcb442-5x5 & 1150909.8 & 1217723.5 & 5.02  & 3419567.0 & 4246700.4 & 1.80  & 786167.2 & 802854.1 & 0.05 \\
		\cline{2-11}          & 36pcb442-6x6 & 1204250.6 & 1279680.5 & 4.07  & -     & -     & -     & 899354.2 & 913260.7 & 0.08 \\
		\cline{2-11}          & 42a280-6x7 & 53430.4 & 57165.5 & 4.07  & -     & -     & -     & 45163.4 & 45660.2 & 0.12 \\
		\cline{2-11}          & 49gil262-7x7 & 41405.4 & 44448.3 & 2.12  & -     & -     & -     & 33206.0 & 33514.1 & 0.15 \\
		\cline{2-11}          & 49lin318-7x7 & 771393.5 & 864248.8 & 2.12  & -     & -     & -     & 591374.9 & 595249.9 & 0.15 \\
		\cline{2-11}          & 9a280-3x3 & 58683.4 & 63126.1 & 2.57  & 99783.6 & 120079.7 & 0.77  & 30443.2 & 31011.6 & 0.02 \\
		\cline{2-11}          & 9gil262-3x3 & 45369.2 & 49783.6 & 2.57  & 76482.4 & 90032.5 & 0.77  & 22158.9 & 23059.2 & 0.02 \\
		\cline{2-11}          & 9lin318-3x3 & 1211011.7 & 1299129.6 & 8.02  & 1672982.0 & 1832321.4 & 1.48  & 730038.2 & 740588.6 & 0.02 \\
		\cline{2-11}          & 9pcb442-3x3 & 1475825.7 & 1620386.4 & 8.02  & 3829863.1 & 4305958.6 & 1.48  & 803179.2 & 821884.9 & 0.02 \\
		\cline{2-11}          & 9pr439-3x3 & 4380949.7 & 4683327.0 & 13.47 & 17129401.8 & 18669027.9 & 1.13  & 1820176.1 & 1881943.0 & 0.03 \\
	    \hline
		\insertTableNotes	  
	\end{longtable}%
	\end{ThreePartTable}	
\end{landscape}


\begin{thebibliography}{10}
\expandafter\ifx\csname url\endcsname\relax
  \def\url#1{\texttt{#1}}\fi
\expandafter\ifx\csname urlprefix\endcsname\relax\def\urlprefix{URL }\fi
\expandafter\ifx\csname href\endcsname\relax
  \def\href#1#2{#2} \def\path#1{#1}\fi

\bibitem{wu_clustered_2015}
B.~Y. Wu, C.-W. Lin, On the clustered {Steiner} tree problem, Journal of
  Combinatorial Optimization 30~(2) (2015) 370--386.

\bibitem{dror_generalized_2000}
M.~Dror, M.~Haouari, J.~Chaouachi,
  \href{http://www.sciencedirect.com/science/article/pii/S0377221799000065}{Generalized
  spanning trees}, European Journal of Operational Research 120~(3) (2000)
  583--592.
\newblock \href {http://dx.doi.org/10.1016/S0377-2217(99)00006-5}
  {\path{doi:10.1016/S0377-2217(99)00006-5}}.
\newline\urlprefix\url{http://www.sciencedirect.com/science/article/pii/S0377221799000065}

\bibitem{wu2014steiner}
B.~Y. Wu, C.-W. Lin, L.-H. Chen, The steiner ratio of the clustered steiner
  tree problem is three, Unpublished manuscript.

\bibitem{mestria2018new}
M.~Mestria, New hybrid heuristic algorithm for the clustered traveling salesman
  problem, Computers \& Industrial Engineering 116 (2018) 1--12.

\bibitem{mestria_grasp_2013}
M.~Mestria, L.~S. Ochi, S.~de~Lima~Martins, {GRASP} with path relinking for the
  symmetric euclidean clustered traveling salesman problem, Computers \&
  Operations Research 40~(12) (2013) 3218--3229.

\bibitem{potvin1996clustered}
J.-Y. Potvin, F.~Guertin, The clustered traveling salesman problem: A genetic
  approach, in: Meta-Heuristics, Springer, 1996, pp. 619--631.

\bibitem{laporte2002some}
G.~Laporte, U.~Palekar, Some applications of the clustered travelling salesman
  problem, Journal of the Operational Research Society 53~(9) (2002) 972--976.

\bibitem{degraeve_optimal_1999}
Z.~Degraeve, L.~Schrage, Optimal integer solutions to industrial cutting stock
  problems, {INFORMS} Journal on Computing 11~(4) (1999) 406--419.

\bibitem{chisman_clustered_1975}
J.~A. Chisman, The clustered traveling salesman problem, Computers \&
  Operations Research 2~(2) (1975) 115--119.

\bibitem{liu_clustering_1999}
C.-M. Liu, Clustering techniques for stock location and order-picking in a
  distribution center, Computers \& Operations Research 26~(10) (1999)
  989--1002.

\bibitem{weintraub_emergency_1999}
A.~Weintraub, J.~Aboud, C.~Fernandez, G.~Laporte, E.~Ramirez, An emergency
  vehicle dispatching system for an electric utility in chile, Journal of the
  Operational Research Society 50~(7) (1999) 690--696.

\bibitem{balakrishnan_scheduling_1992}
N.~Balakrishnan, A.~Lucena, R.~T. Wong, Scheduling examinations to reduce
  second-order conflicts, Computers \& Operations Research 19~(5) (1992)
  353--361.

\bibitem{laporte1998tiling}
G.~Laporte, F.~Semet, V.~Dadeshidze, L.~Olsson, A tiling and routing heuristic
  for the screening of cytological samples, Journal of the Operational Research
  Society 49~(12) (1998) 1233--1238.

\bibitem{laporte_applications_2002}
G.~Laporte, U.~Palekar, Some applications of the clustered travelling salesman
  problem, Journal of the Operational Research Society 53~(9) (2002) 972--976.

\bibitem{back_evolutionary_1996}
T.~Back, Evolutionary algorithms in theory and practice: evolution strategies,
  evolutionary programming, genetic algorithms, Oxford university press, 1996.

\bibitem{ding2007two}
C.~Ding, Y.~Cheng, M.~He, Two-level genetic algorithm for clustered traveling
  salesman problem with application in large-scale tsps, Tsinghua Science and
  technology 12~(4) (2007) 459--465.

\bibitem{pop2018two}
P.~Pop, O.~Matei, C.~Pintea, A two-level diploid genetic based algorithm for
  solving the family traveling salesman problem, in: Proceedings of the Genetic
  and Evolutionary Computation Conference, ACM, 2018, pp. 340--346.

\bibitem{pop2017hybrid}
P.~Pop, M.~Oliviu, C.~Sabo, A hybrid diploid genetic based algorithm for
  solving the generalized traveling salesman problem, in: International
  Conference on Hybrid Artificial Intelligence Systems, Springer, 2017, pp.
  149--160.

\bibitem{lin_minimum_2016}
C.-W. Lin, B.~Y. Wu, On the minimum routing cost clustered tree problem,
  Journal of Combinatorial Optimization (2016) 1--16.

\bibitem{demidio_clustered_2016}
M.~D'Emidio, L.~Forlizzi, D.~Frigioni, S.~Leucci, G.~Proietti, On the
  {Clustered} {Shortest}-{Path} {Tree} {Problem}., in: {ICTCS}, 2016, pp.
  263--268.

\bibitem{agoston_eiben_2003}
E.~E. Agoston, {Introduction} to {Evolutionary} {Computing}, Berlin,
  Springer-Verlag, 2003.

\bibitem{goldberg2006genetic}
D.~E. Goldberg, Genetic algorithms, Pearson Education India, 2006.

\bibitem{ThanhPD_TrungTB}
H.~T.~T. Binh, P.~D. Thanh, T.~B. Trung, L.~P. Thao, Effective multifactorial
  evolutionary algorithm for solving the cluster shortest path tree problem,
  in: Evolutionary {Computation} ({CEC}), 2018 {IEEE} {Congress} on, IEEE,
  2018, pp. 819--826.

\bibitem{ThanhPD_DungDA}
P.~D. Thanh, D.~A. Dung, T.~N. Tien, H.~T.~T. Binh, An effective representation
  scheme in multifactorial evolutionary algorithm for solving cluster
  shortest-path tree problem, in: Evolutionary {Computation} ({CEC}), 2018
  {IEEE} {Congress} on, IEEE, 2018, pp. 811--818.

\bibitem{thompson2007dandelion}
E.~Thompson, T.~Paulden, D.~K. Smith, The dandelion code: A new coding of
  spanning trees for genetic algorithms, IEEE Transactions on Evolutionary
  Computation 11~(1) (2007) 91--100.

\bibitem{perfecto2016dandelion}
C.~Perfecto, M.~N. Bilbao, J.~Del~Ser, A.~Ferro, S.~Salcedo-Sanz,
  Dandelion-encoded harmony search heuristics for opportunistic traffic
  offloading in synthetically modeled mobile networks, in: Harmony Search
  Algorithm, Springer, 2016, pp. 133--145.

\bibitem{julstrom2005blob}
B.~A. Julstrom, The blob code is competitive with edge-sets in genetic
  algorithms for the minimum routing cost spanning tree problem, in:
  Proceedings of the 7th annual conference on Genetic and evolutionary
  computation, ACM, 2005, pp. 585--590.

\bibitem{palmer_representing_1994}
C.~Palmer, A.~Kershenbaum,
  \href{http://ieeexplore.ieee.org/document/349921/}{Representing trees in
  genetic algorithms}, IEEE, Orlando, FL, USA, 1994, pp. 379--384.
\newblock \href {http://dx.doi.org/10.1109/ICEC.1994.349921}
  {\path{doi:10.1109/ICEC.1994.349921}}.
\newline\urlprefix\url{http://ieeexplore.ieee.org/document/349921/}

\bibitem{paulden_recent_2006}
T.~Paulden, D.~Smith,
  \href{http://ieeexplore.ieee.org/document/1688567/}{Recent {Advances} in the
  {Study} of the {Dandelion} {Code}, {Happy} {Code}, and {Blob} {Code}
  {Spanning} {Tree} {Representations}}, IEEE, Vancouver, BC, Canada, 2006, pp.
  2111--2118.
\newblock \href {http://dx.doi.org/10.1109/CEC.2006.1688567}
  {\path{doi:10.1109/CEC.2006.1688567}}.
\newline\urlprefix\url{http://ieeexplore.ieee.org/document/1688567/}

\bibitem{pop2018novel}
P.~C. Pop, L.~Fuksz, A.~H. Marc, C.~Sabo, A novel two-level optimization
  approach for clustered vehicle routing problem, Computers \& Industrial
  Engineering 115 (2018) 304--318.

\bibitem{pop2018twospanningtree}
P.~C. Pop, O.~Matei, C.~Sabo, A.~Petrovan, A two-level solution approach for
  solving the generalized minimum spanning tree problem, European Journal of
  Operational Research 265~(2) (2018) 478--487.

\bibitem{deb2014evolutionary}
K.~Deb, A.~Sinha, Evolutionary bilevel optimization (ebo), in: Proceedings of
  the Companion Publication of the 2014 Annual Conference on Genetic and
  Evolutionary Computation, ACM, 2014, pp. 857--876.

\bibitem{sinha2017evolutionary}
A.~Sinha, P.~Malo, K.~Deb, Evolutionary bilevel optimization: An introduction
  and recent advances, in: Recent Advances in Evolutionary Multi-objective
  Optimization, Springer, 2017, pp. 71--103.

\bibitem{shu2012improved}
W.~Shu-Xi, The improved dijkstra's shortest path algorithm and its application,
  Procedia Engineering 29 (2012) 1186--1190.

\bibitem{xu2007improved}
M.~Xu, Y.~Liu, Q.~Huang, Y.~Zhang, G.~Luan, An improved dijkstra’s shortest
  path algorithm for sparse network, Applied Mathematics and Computation
  185~(1) (2007) 247--254.

\bibitem{johnson1973note}
D.~B. Johnson, A note on dijkstra's shortest path algorithm, Journal of the ACM
  (JACM) 20~(3) (1973) 385--388.

\bibitem{dijkstra1959note}
E.~W. Dijkstra, A note on two problems in connexion with graphs, Numerische
  mathematik 1~(1) (1959) 269--271.

\bibitem{helsgaun_solving_2011}
K.~Helsgaun, Solving the {Clustered} {Traveling} {Salesman} {Problem} {Using}
  the {Lin}-{Kernighan}-{Helsgaun} {Algorithm}, Computer Science Research
  Report~(142) (2011) 1--16.

\bibitem{Pham_Dinh_Thanh_2018_Instances}
P.~D. Thanh, Clu{SPT} instances, Mendeley Data v2, 2018.
\newblock \href {http://dx.doi.org/http://dx.doi.org/10.17632/b4gcgybvt6.2}
  {\path{doi:http://dx.doi.org/10.17632/b4gcgybvt6.2}}.

\end{thebibliography}
\end{document}